\def\year{2020}\relax
\documentclass[letterpaper]{article} 
\usepackage{aaai20}  
\usepackage{times}  
\usepackage{helvet} 
\usepackage{courier}  
\usepackage[hyphens]{url}  
\usepackage{graphicx} 
\usepackage{graphicx}  
\usepackage{float}
\usepackage[pagebackref=true,breaklinks=true,colorlinks,citecolor=green,urlcolor=magenta,bookmarks=false]{hyperref}
\usepackage{dirtytalk}
\usepackage{caption}
\usepackage{subcaption}
\usepackage{epsfig}
\usepackage{ dsfont }
\usepackage{amsmath}
\usepackage{amssymb}
\usepackage{booktabs}

\title{\fontsize{13.5}{14}\selectfont One Framework to Register Them All: PointNet Encoding for Point Cloud Alignment}

\author{\fontsize{9}{13}\selectfont \text{Vinit Sarode}\textsuperscript{1}\thanks{equal contribution},
\text{ Xueqian Li}\textsuperscript{1*},
\text{ Hunter Goforth}\textsuperscript{3},
\text{ Yasuhiro Aoki}\textsuperscript{2},
\text{ Animesh Dhagat},\textsuperscript{1}\\
\fontsize{9}{13}\textbf{ Rangaprasad Arun Srivatsan}\textsuperscript{4},
\textbf{ Simon Lucey}\textsuperscript{1,3},
\textbf{ Howie Choset}\textsuperscript{1}\\
$^\text{1}$Carnegie Mellon University
\and
$^\text{2}$Fujitsu Laboratories Ltd.
\and
$^\text{3}$Argo AI.
\and
$^\text{4}$Apple.\\
{\tt \{vsarode, xueqianl, adhagat\}@andrew.cmu.edu },
{\tt  aoki-yasuhiro@fujitsu.com}
\\
{\tt \{hgoforth, slucey, choset\}@cs.cmu.edu }
,
{\tt  aruns@apple.com}}

\begin{document}

\maketitle
\begin{abstract}
PointNet has recently emerged as a popular representation for unstructured point cloud data, allowing application of deep learning to tasks such as object detection, segmentation and shape completion. However, recent works in literature have shown the sensitivity of the PointNet representation to pose misalignment. This paper presents a novel framework that uses PointNet encoding to align point clouds and perform registration for applications such as 3D reconstruction, tracking and pose estimation. We develop a framework that compares PointNet features of template and source point clouds to find the transformation that aligns them accurately. In doing so, we avoid computationally expensive correspondence finding steps, that are central to popular registration methods such as ICP and its variants. Depending on the prior information about the shape of the object formed by the point clouds, our framework can produce approaches that are shape specific or general to unseen shapes. Our framework produces approaches that are robust to noise and initial misalignment in data and work robustly with sparse as well as partial point clouds. We perform extensive simulation and real-world experiments to validate the efficacy of our approach and compare the performance with state-of-art approaches. Code is available at
~\href{https://github.com/vinits5/pointnet-registration-framework.git}{\textit{https://github.com/vinits5/pointnet-registration-framework}}.
\end{abstract}

\begin{figure}[t!]
    \centering
    \includegraphics[width=0.88\columnwidth]{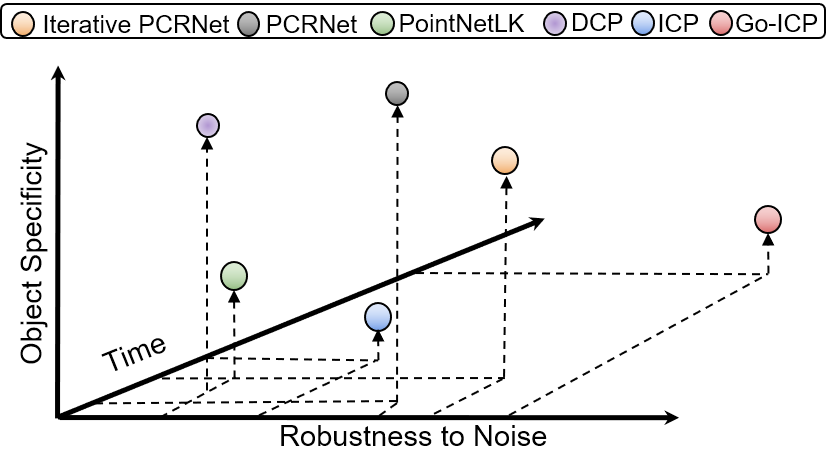}
    \caption{
    `No free lunch' in point cloud registration. Comparison of different registration methods based on their robustness to noise and computation time with respect to object specificity. 
    The iterative version of point cloud registration network (PCRNet) exploits object specificity to produce accurate results. The PCRNet without iterations is computationally faster and more robust to noise, but compromises a little on accuracy. 
    Deep closest point~\protect\cite{wang2019deep} is also computationally fast, but not as robust to noise. 
    While  PointNetLK~\protect\cite{aoki2019pointnetlk} exhibits good generalizability to unseen objects. ICP~\protect\cite{INTRO:ICP} is object-shape agnostic and slow for large point clouds, while Go-ICP~\protect\cite{RW:GO_ICP} is computationally expensive.
    }
    \label{fig:firstfig}
    \vspace{-3mm}
\end{figure}

\section{Introduction}
3D point clouds are ubiquitous today, thanks to the development of low-cost and reliable lidar, stereo cameras and structured light sensors. As a result there has been a growing interest in developing algorithms for performing classification, segmentation, tracking, mapping, etc. directly using point clouds. However, the inherent lack of structure presents difficulties in using point clouds directly in deep learning architectures. Recent developments such as PointNet~\cite{qi2017pointnet} and its variants~\cite{qi2017pointnet++} have been instrumental in overcoming some of these difficulties, resulting in state-of-the-art methods for object detection and segmentation tasks~\cite{RW:frustum,RW:Wentao}.

Prior works~\cite{RW:Wentao,aoki2019pointnetlk} have observed that robust performance of PointNet requires minimal misalignment of the point clouds with respect to a canonical coordinate frame. While this is present in synthetic datasets such as \emph{ModelNet40}~\cite{wu20153d}, real world data is seldom aligned to some canonical coordinate frame. Inspired by recent works on iterative transformer network (IT-Net)~\cite{RW:Wentao} and PointNetLK~\cite{aoki2019pointnetlk}, this work introduces a framework for estimating the misalignment between two point clouds using PointNet as an encoding function. It is worth noting that our approach can directly process point clouds for the task of registration, without the need for hand crafted features~\cite{rusu2009fast,gelfand2005robust}, voxelization~\cite{maturana2015voxnet,gojcic2019perfect} or mesh generation~\cite{wang2018dynamic}. Our framework provides approaches that can utilize prior knowledge of the shape of the object being registered, to robustly deal with noise, sparse measurements and incomplete data. Our framework also provides additional context for PointNetLK (see Sec.~\ref{sec:one-shot registration network} for more details). It is worth emphasizing, that we do not propose a single registration technique that outperforms all state-of-the-art methods. Instead we propose a framework that generates a number of registration approaches, including some that already exist in literature. The performance of the various approaches produced by our framework depend on factors such as the prior information about the shape of the object, noise in the measurements and computation time (see Fig.~\ref{fig:firstfig}).

Our approach uses PointNet in a siamese architecture to encode the shape information of a template and a source point cloud as feature vectors, and estimates the pose that aligns these two features using data driven techniques. The pose estimation from the features are carried out either using a number of fully connected (FC) layers or using classical alignment techniques such as Lucas-Kanade (LK) algorithm~\cite{lucas1981iterative,baker2004lucas}. The LK algorithm results in good generalizability, but is not robust to noise. The FC layers are robust to noise, but not generalizable to shapes unseen during training.

Using shape-specific prior information in the training phase allows us to be robust to noise in the data, compared to shape agnostic methods such as  iterative closest point (ICP)~\cite{INTRO:ICP} and its variants~\cite{rusinkiewicz2001efficient}. Unlike ICP, our approach does not require costly closest point correspondence computations, resulting in improved computational efficiency and robustness to noise. Further, the approach is fully differentiable which allows for easy integration with other deep networks and can be run directly on GPU without need for any CPU computations.

Our contributions are (1) presenting a novel framework for point cloud alignment which utilize PointNet representation for effective correspondence-free registration, and (2) a thorough experimental validation of our approaches including comparison against popular and state-of-the-art registration methods (both conventional and learning-based approaches), on both simulated and real-world data.

\section{Related Work}
\paragraph{Classical registration.}
Iterative Closest Point (ICP) remains one of the most popular techniques for point cloud registration, as it is straightforward to implement and produces adequate results in many scenarios~\cite{INTRO:ICP}. Extensions of ICP have increased computational efficiency~\cite{rusinkiewicz2001efficient,srivatsan2019registration} and improved accuracy~\cite{RW:GO_ICP}. However, nearly all ICP variants rely on explicit computation of closest point correspondences, a process which scales poorly with the number of points. Additionally, ICP is not differentiable (due to the requirement to find discrete point correspondences) and thus cannot be integrated into end-to-end deep learning pipelines, inhibiting the ability to apply learned descriptors for alignment.

\textit{Interest point} methods compute and compare local descriptors to estimate alignment~\cite{gelfand2005robust,guo20143d}. These methods have the advantage of being computationally favorable, however, their use is often limited to point cloud data having identifiable and unique features which are persistent between point clouds that are being registered~\cite{makadia2006fully,ovsjanikov2010one,rusu2009fast}.


\textit{Globally optimal} methods~\cite{izatt2017,maron2016point} seek to find optimal solutions which cannot reliably be found with iterative techniques such as ICP. Unfortunately, these techniques are characterized by extended computation times, which largely precludes their use in applications requiring real-time speed. A representative example which we use as a baseline is Go-ICP~\cite{RW:GO_ICP}, a technique using branch-and-bound optimization.

\paragraph{PointNet.}
PointNet is the first deep neural network which processes point clouds directly~\cite{qi2017pointnet}, as opposed to alternative representations such as 2D image projections of objects~\cite{xiang2017posecnn,RW:Semantic,RW:CAD}, voxel representations~\cite{maturana2015voxnet,wu20153d,zhou2018voxelnet} or graph representations~\cite{wang2018dynamic}. Within larger network architectures, PointNet has proven to be useful for tasks including classification, semantic segmentation, object detection~\cite{RW:frustum}, flow estimation~\cite{liu2019flownet3d}, and completion of partial point clouds~\cite{yuan2018pcn}. An extension to PointNet for estimating local feature descriptors is described in~\cite{qi2017pointnet++}. \citeauthor{RW:Wentao} introduced iterative transformer network (IT-Net) which uses PointNet to estimate a canonical orientation of point clouds to increase classification and segmentation accuracy. Global descriptors from PointNet are used in~\cite{angelina2018pointnetvlad} for place recognition from 3D data.

\begin{figure}[t!]
    \centering
    \includegraphics[width=1\linewidth]{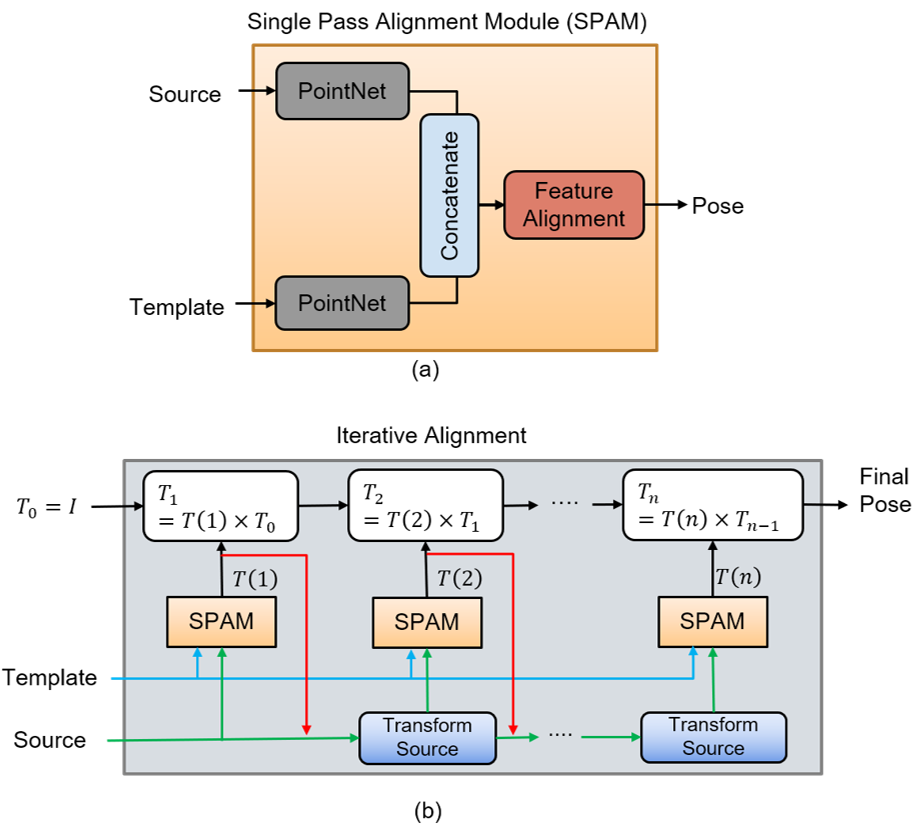}
    \caption{(a) Single Pass Alignment Module (SPAM): This module estimates the pose in a single pass, from a source and template point cloud. PointNet is used in a siamese architecture to extract a global feature vectors from both source and template. The point cloud alignment problem is posed in terms of alignment of the features. This circumvents the problem of finding explicit point correspondences. Different choices for the feature alignment algorithms gives rise to different approaches with their inherent advantages and disadvantages. (b) Iterative Alignment: The output of SPAM can be iteratively improved. After each iteration, the source point cloud is transformed using the pose estimated from the previous iteration. After performing $n$ iterations, the poses from each iteration are combined to find the overall transformation.
}
    \label{fig:one_shot_registration}
    \vspace{-3mm}
\end{figure}

\paragraph{Learned registration.}
Early learning-based approaches use a combination of hand-crafted and learned features and learned map sets for the task of point cloud registration~\cite{vongkulbhisal2018inverse}.  Deep auto-encoders are used to extract local descriptors for registration of large outdoor point clouds in~\cite{elbaz20173d}. \citeauthor{yew20183dfeat} introduced a network which learns both interest point detection and descriptor computation, for a descriptor-matching registration approach. More recently \citeauthor{lu2019deepicp} developed Deep-ICP, an approach that learns correspondences between point clouds and then uses an SVD to align the points similar to ICP~\cite{lu2019deepicp}. A major shortcoming of all these approaches is they do not typically scale well with increase in the number of points being registered, and lack generalization due to the feature vector and registration maps both being learned.

PointNetLK~\cite{aoki2019pointnetlk}, which performs registration of arbitrary point clouds by minimizing the distance between the fixed-length, global descriptors produced by PointNet, is the most closely related to our work and serves as a baseline. Another work that comes close to ours is the siamese network used by \citeauthor{zhou2019continuity} to estimate the orientation between two point clouds~\cite{zhou2019continuity}. As an alternate to PointNet encoding, \citeauthor{wang2018dynamic} perform convolution operations on the edges that connect neighboring point pairs, by using a local neighborhood graph~\cite{wang2018dynamic}. They introduced a network called Deep Closest Point (DCP), which uses this graph to perform point cloud alignment~\cite{wang2019deep}. We also present comparisons to this method in this work.

In addition to the above mentioned methods, there are several learning-based approaches that perform alignment with RGB-D data~\cite{pais20193dregnet,insafutdinov2018unsupervised,li2018deepim,wang2019densefusion}. Since we use only point cloud data and no associated RGB information in this work, we restrict our comparisons to methods that use only point clouds for alignment.  

\section{Method}
\label{section:method}
Point clouds are highly unstructured with ambiguities in the order permutations. While performing classification using PointNet, a symmetric pooling function such as max pool is used to afford invariance to input permutation (see~\cite{qi2017pointnet} for more details). The output vector of the symmetry function is referred to as a global feature vector. We denote the template point cloud \textbf{P}$_T$ and source \textbf{P}$_S$, and the PointNet function $\phi$. Since the global feature vectors contain the information about the geometry as well as the orientation of the point clouds, the transformation between two point clouds can be obtained by comparing the feature vectors. In other words, we calculate the rigid-body transformation $\textbf{T}\in SE(3)$, that minimizes the difference between $\phi$(\textbf{P}$_S$) and $\phi$(\textbf{P}$_T$). 

\subsection{Single Pass Alignment Module}
This section introduces an alignment module that is central to the framework (see Fig.~\ref{fig:one_shot_registration}(a)). This module takes as input a point cloud data obtained from a sensor, which is referred to as the \textit{source} and a point cloud corresponding to the known model of the object to be registered, which is referred to as the \textit{template}. Both source \textbf{P}$_S$ and template \textbf{P}$_T$ are given as input to a PointNet module. The PointNet internally hasseveral layers of multi-layered perceptrons (MLPs),  which are arranged in a Siamese architecture~\cite{held2016learning}.  A symmetric max-pooling function is used to find the global feature vectors $\phi$(\textbf{P}$_S$) and $\phi$(\textbf{P}$_T$). Weights are shared between the MLPs used for source and template. The model consists of five MLPs having size 64, 64, 64, 128, 1024. The global features are concatenated and given as an input to a feature alignment module. This module either uses classical alignment techniques of uses data driven techniques to learn the alignment between the features. 

\subsection{Iterative Alignment}
Inspired by iterative schemes for alignment problem such as~\cite{baker2004lucas,INTRO:ICP,li2018deepim}, etc., we use SPAM to refine the pose estimate after each iteration and obtain an accurate alignment between the source and template point clouds (see Fig.~\ref{fig:one_shot_registration}(b)). 

In the first iteration, original source and template point clouds are given to SPAM which predicts an initial alignment $\textbf{T}(1)\in SE(3)$ between them. For the next iteration, $\textbf{T}(1)$ is applied to the source point cloud and then the transformed source and the original template point clouds are given as input to the SPAM. After performing $n$ iterations, we find the overall transformation between the original source and template point clouds by combining all the poses in each iteration:
\mbox{$\textbf{T} =\textbf{T}(n)\times\textbf{T}(n-1)\times \cdots\times\textbf{T}(1)$}.

Depending on the choice of feature alignment algorithm, number of iterations, and choice of loss functions several different approaches can be produced by this framework. We explain three approaches in the next section, namely: PCRNet, i-PCRNet and PointNetLK.

\paragraph{PCRNet}
\label{sec:one-shot registration network}
This section introduces the point cloud registration network (PCRNet) architecture. The PCRNet is a single pass pose estimator, which uses data driven techniques to align the PointNet features. Five fully connected layers of size 1024,1024,512,512,256 are used along with an output layer of the dimension of the parameterization chosen for the pose. We tried using lesser number of FC layers, but the performance of the network was poor.

The transformation \textbf{T} which aligns $\phi$(\textbf{P}$_S$) and $\phi$(\textbf{P}$_T$) is estimated with a single forward pass, or single-shot, through the network. The single-shot design lends itself particularly well to high-speed applications, which will be discussed further in Sec.~\ref{sec:results}.

\paragraph{Iterative PCRNet}
In this section, we introduce iterative PCRNet (i-PCRNet). The i-PCRNet uses a modified form of PCRNet as the single pass alignment module in Fig.~\ref{fig:one_shot_registration}(b). We retain the structure of PCRNet but modify the number of layers. The fully connected layers have three hidden layers with size 1024, 512, 256. Also, there is an additional dropout layer before the output layer, to avoid overfitting. We empirically observe that introducing iterations allows us to use lesser number of hidden layers compared to PCRNet, and yet obtain robust performance. 

\paragraph{PointNetLK}
The PointNetLK was introduced by~\citeauthor{aoki2019pointnetlk}. We observe that PointNetLK is just another special case of our framework. If we were to use an inverse compositional Lucas-Kanade algorithm~\cite{baker2004lucas} for aligning the features, while still performing the iterations similar to i-PCRNet, the resulting implementation is PointNetLK.

\paragraph{Pose Parameterization}
The transformation $\textbf{T}$, can be parameterized in a number of different ways. We tried several parameterizations namely Cartesian coordinates and unit quaternions, Euler angles, twist coordinates, 6D continuous parameters~\cite{zhou2019continuity}, and 12D parameters~\cite{pais20193dregnet}. Contrary to the observations of \citeauthor{zhou2019continuity}, we do not observe any significant improvement in using one over the other.  
\paragraph{Loss Function} There are several choices for loss functions that can be used to train the networks. We considered three options, Frobenius norm~\cite{aoki2019pointnetlk}, EMD~\cite{RW:Wentao} and chamfer distance~\cite{fan2017point}. From Fig.~\ref{fig:iterations_vs_error}(a), we observe that while all three loss functions perform well, CD slightly outperforms the other two.

\paragraph{Training}
In this work, we use \emph{ModelNet40} dataset~\cite{wu20153d} to train the network. This dataset contains CAD models of 40 different object categories. We uniformly sample points based on face area and then used farthest point algorithm~\cite{eldar1997farthest} to get a complete point cloud.
We train the networks with three different types of datasets as following -- (1) Multiple categories of objects and multiple models from each category, (2) Multiple models of a specific category, (3) A single model from a specific category. We choose these 3 cases to showcase the performance of the PointNet-based approaches on data with differing levels of object-specificity.

We train the i-PCRNet with 8 iterations during training, observing that more than 8 produced little improvement to results. In some experiments the training data was corrupted with Gaussian noise, which is discussed in detail in Sec.~\ref{sec:Gaussian_Noise}. The networks are trained for 300 epochs, using a learning rate of $10^{-1}$ with an exponential decay rate of 0.7 after every $3\times10^{6}$ steps and batch size 32. The network parameters are updated with Adam Optimizer on a single NVIDIA GeForce GTX 1070 GPU and a Intel Core i7 CPU at 4.0GHz.

\begin{figure}[t!]
\centering
    \begin{subfigure}[t]{0.21\textwidth}
        \centering
        \includegraphics[width=\linewidth]{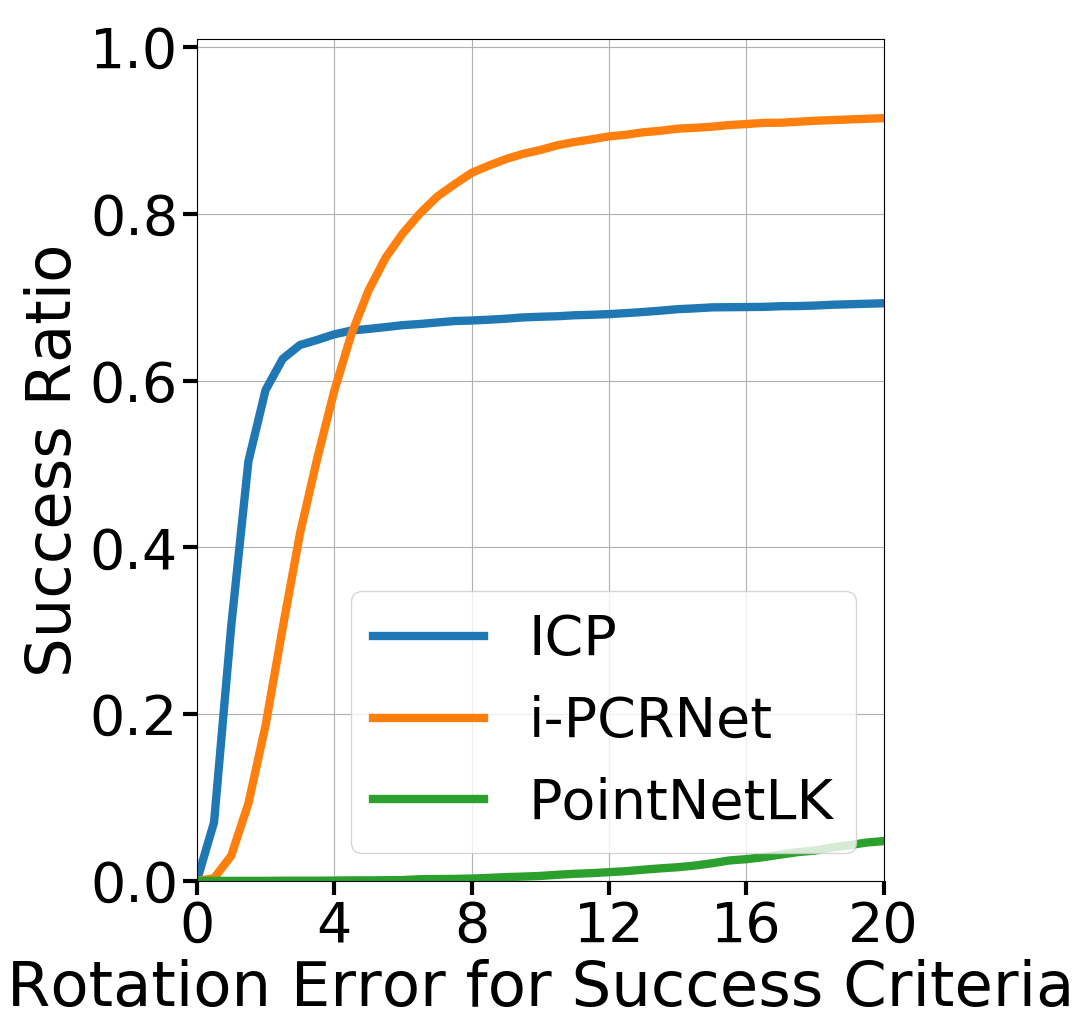}
        \caption{Training and testing: Multiple object categories with noise.}
        \label{fig:multi_catg_noise}
    \end{subfigure}
    ~
    \begin{subfigure}[t]{0.21\textwidth}
        \centering
        \includegraphics[width=\linewidth]{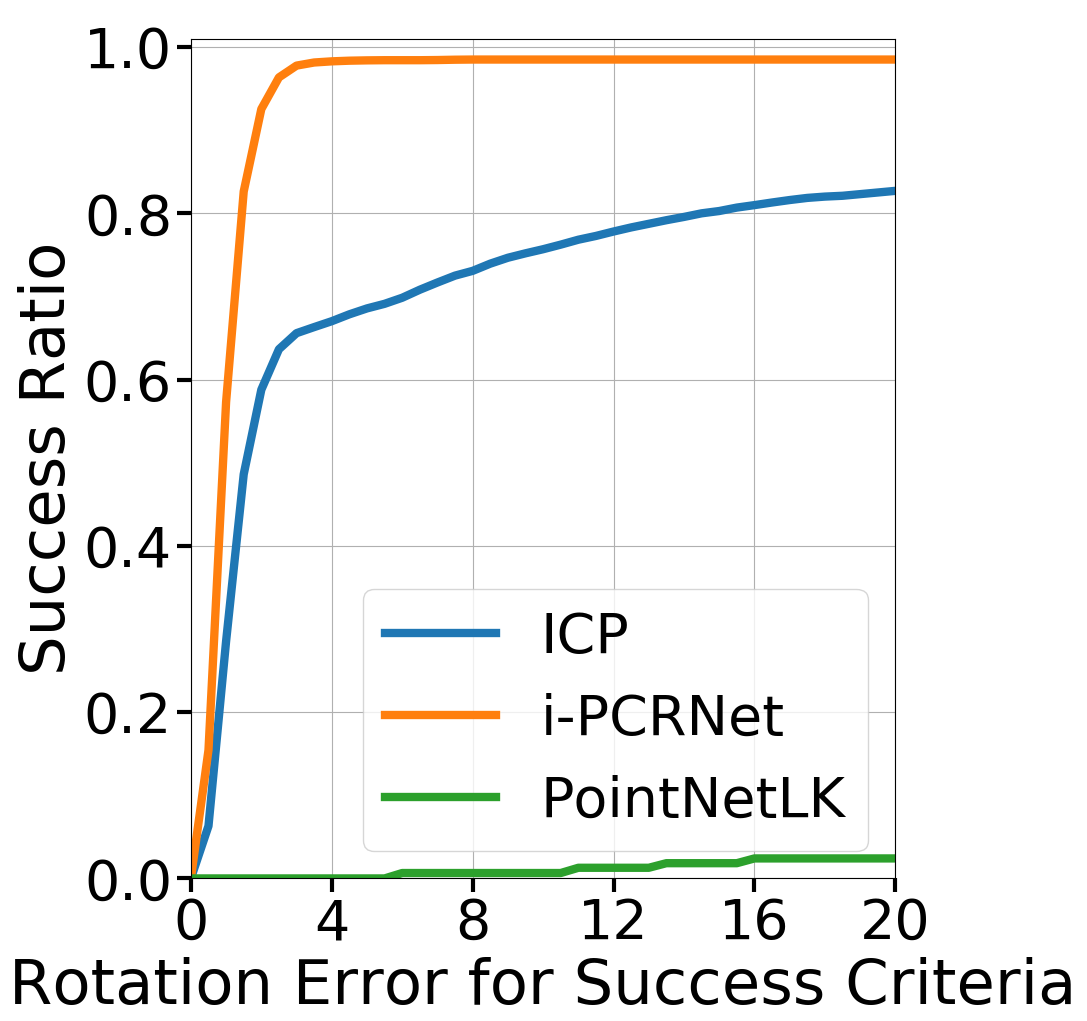}
        \caption{Training and testing: Single category with noise.}
        \label{fig:one_catg_noise}
    \end{subfigure}
    ~
    \begin{subfigure}[t]{0.21\textwidth}
        \centering
         \includegraphics[width=\linewidth]{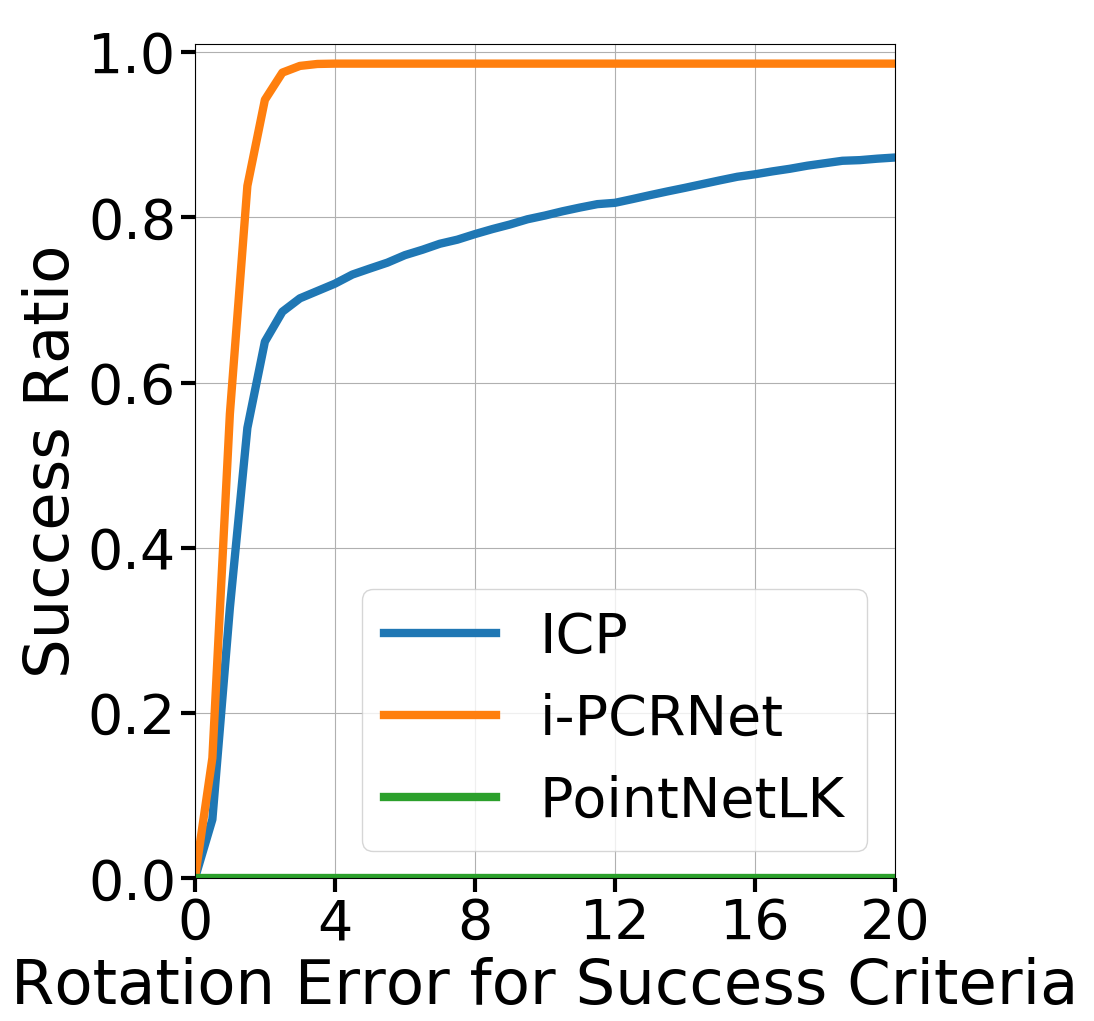}
         \caption{Training and testing: Single model with noise.}
         \label{fig:one_model_noise}
    \end{subfigure}
    ~
    \begin{subfigure}[t]{0.24\textwidth}
        \centering
         \includegraphics[width=0.9\linewidth]{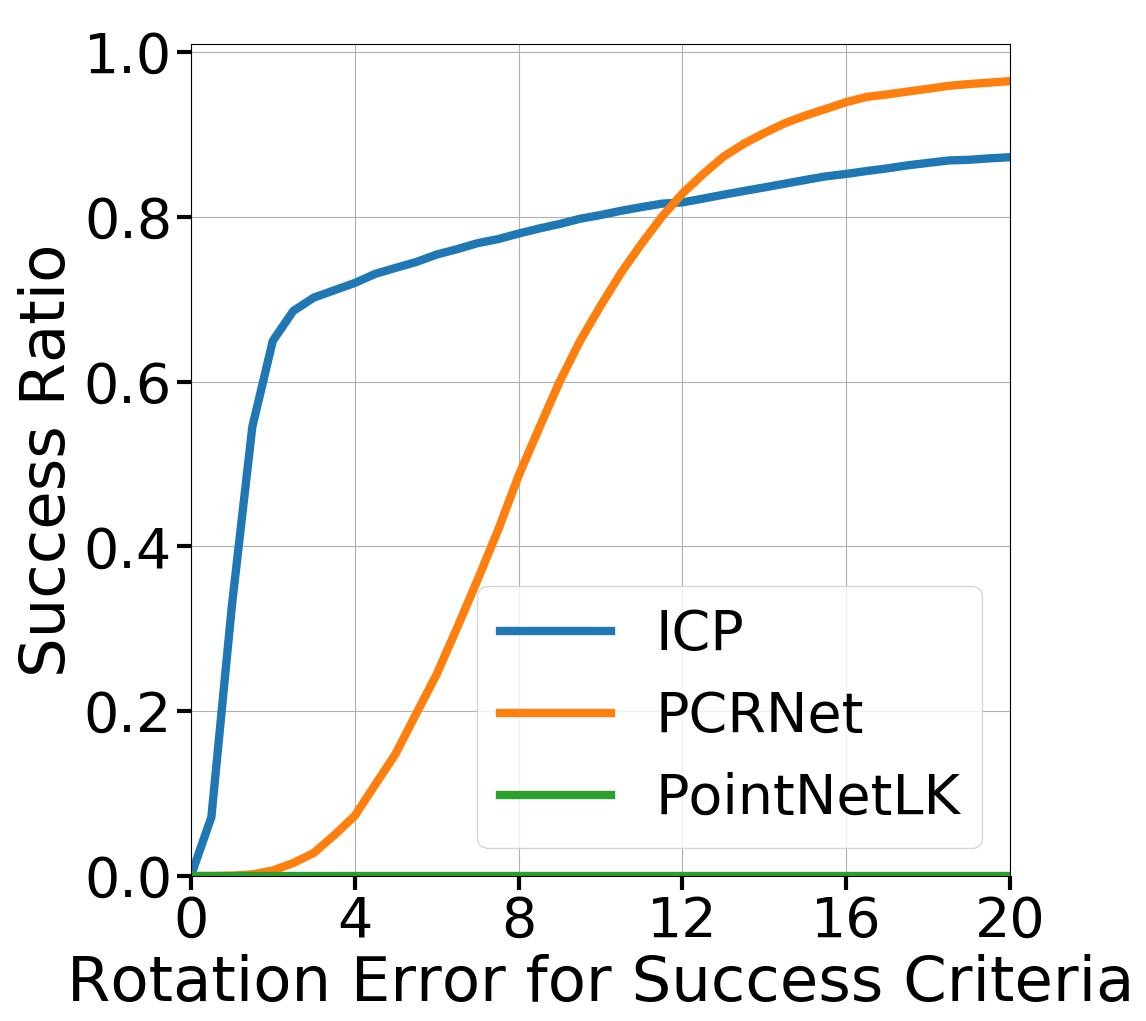}
         \caption{Training: model without noise. Testing: model with noise.}
         \label{fig:one_model_noise_siam}
    \end{subfigure}
    \caption{Results for Section~\ref{sec:Gaussian_Noise}. The $y$-axis is the ratio of experiments that are successful and the $x$-axis shows value of the maximum rotation error that qualifies the estimation to be a success. (a), (b) and (c) shows results for comparisons of i-PCRNet with ICP and PointNetLK using three different types of datasets. We observe superior performance of i-PCRNet as our network has more model/category specific information. (d) PCRNet which has not seen noise during training but tested with noisy data also shows good performance and is faster than ICP and PointNetLK.}
    \label{fig:with_noise}
    \vspace{-3mm}
\end{figure}

\begin{figure*}[t!]
    \centering
    \begin{subfigure}{0.26\textwidth}
        \centering
        \includegraphics[width=\linewidth]{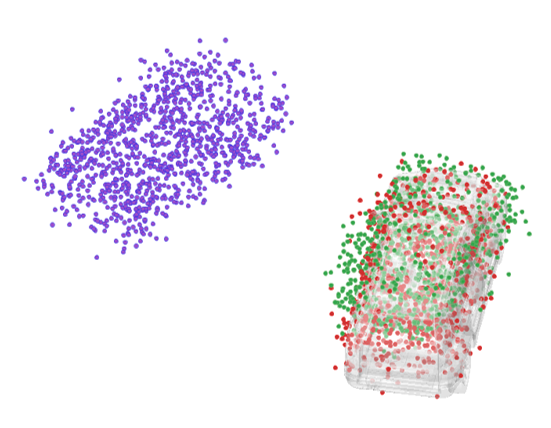}
        \caption{Trained on one car \\ i-PCRNet: Rot error = $2.14^\circ$,\\ Trans error = $0.0056$ units.}
        \label{fig:one_model_seen}
    \end{subfigure}
    ~
    \begin{subfigure}{0.26\textwidth}
        \centering
         \includegraphics[width=\linewidth]{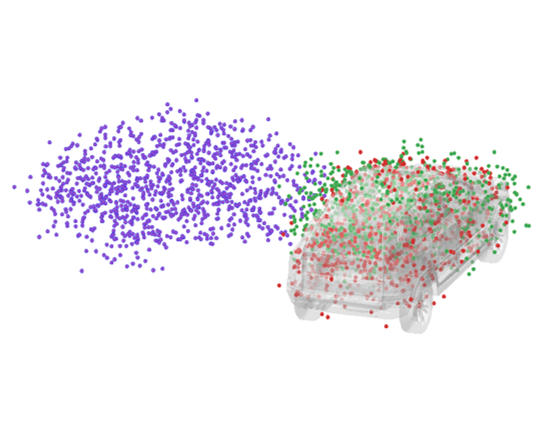}
         \caption{Trained on multiple cars \\ i-PCRNet: Rot error = $2.14^\circ$,\\ Trans error = $0.0056$ units.}
         \label{fig:multi_models_seen}
    \end{subfigure}
    ~
    \begin{subfigure}{0.26\textwidth}
        \centering
        \includegraphics[width=\linewidth]{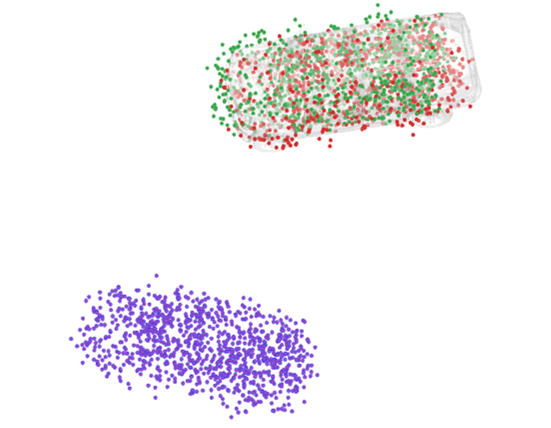}
        \caption{Trained on multiple categories \\ i-PCRNet: Rot error = $3.07^\circ$, \\ Trans error = $0.0107$ units.}
        \label{fig:multi_catgs_seen}
    \end{subfigure}
    ~
    \begin{subfigure}{0.26\textwidth}
        \centering
         \includegraphics[width=0.6\linewidth]{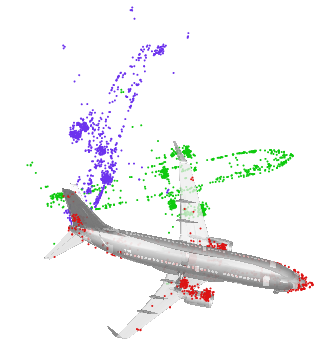}
         \caption{Trained on multiple categories \\ i-PCRNet: Rot error = $0.34^\circ$,\\ Trans error = $0.0048$ units. \\ ICP: Rot error = $43.62^\circ$, \\ Trans error = $0.2564$ units.}
         \label{fig:multi_catgs_seen_1}
    \end{subfigure}
    ~
    \begin{subfigure}{0.26\textwidth}
        \centering
         \includegraphics[width=0.8\linewidth]{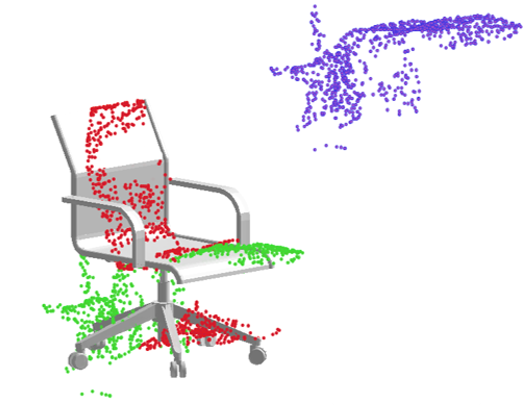}
         \caption{Trained on multiple categories \\ Registration of chair point cloud taken from Stanford \emph{S3DIS} indoor dataset~\cite{armeni_cvpr16}.         }
         \label{fig:multi_catgs_seen_2}
    \end{subfigure}
    ~
    \begin{subfigure}{0.26\textwidth}
        \centering
         \includegraphics[width=1\linewidth]{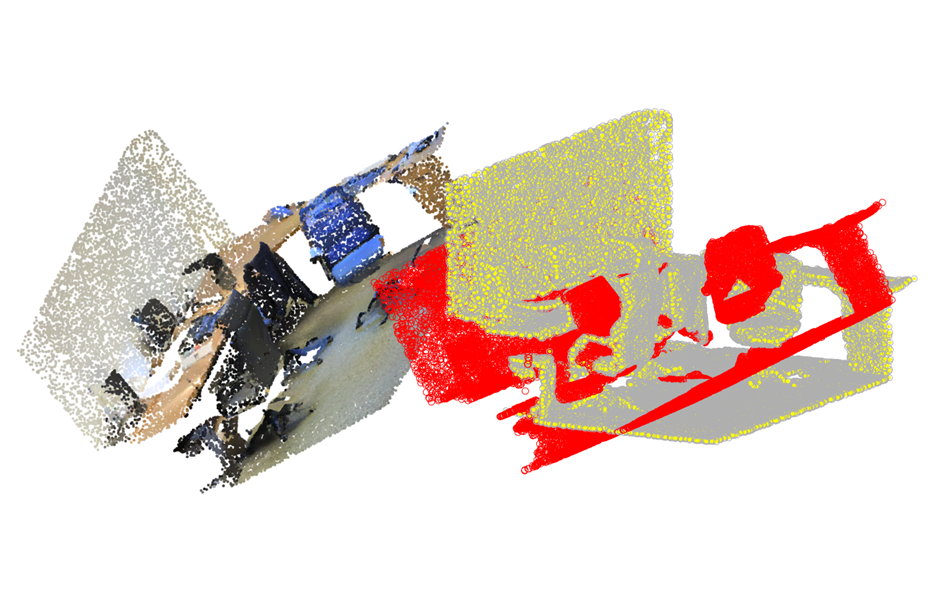}
         \caption{Trained on multiple categories \\ i-PCRNet: Rot error = $26.90^\circ$,\\ Trans error = $0.8157$ units. \\ PointNetLK: Rot error = $0^\circ$,\\ Trans error = $0$ units.}
         \label{fig:multi_catgs_unseen_1}
    \end{subfigure}
    \caption{Qualitative results for Section \ref{sec:results}. For each example, template is shown by a grey rendered CAD model, purple points show initial position of source and red points show converged results of i-PCRNet trained on data with noise and green points show results of ICP. (d) shows a result with sparse point cloud, (e) shows a result with partial point cloud, and (f) shows a result of unseen category for PointNetLK with yellow points and i-PCRNet with red points. For (a) - (e), where the test data has some representation in training, i-PCRNet performs better. On the other hand, in the case of (f) where the test data is obtained from an RGBD scan and is unseen during training, PointNetLK performs better. }
    \label{fig:example_registration}
    \vspace{-3mm}
\end{figure*}

\section{Results} \label{sec:results}
In this section, we compare performance of our networks on test data with multiple object categories, a specific object category, a specific object from training dataset and objects unseen in training. We use models from \emph{ModelNet40} dataset~\cite{wu20153d} for the following experiments. Template point clouds are normalized into a unit box and then their mean is shifted to origin. We randomly choose 5070 transformations with Euler angles in the range of $[-45^\circ, 45^\circ]$ and translation values in the range of [-1, 1] units. We apply these rigid transformations on the template point clouds to generate the source point clouds. We allow a maximum of 20 iterations for both i-PCRNet and PointNetLK while performing tests, while the maximum iterations for ICP was chosen as 100. In addition to maximum iterations, we also use the convergence criteria $\left\| \textbf{T}_{i}\textbf{T}_{i-1}^{-1}-\textbf{I} \right\|_{F} < \epsilon$,where $\textbf{T}_{i}, \textbf{T}_{i-1} \in SE(3)$ are the transformations predicted in current and previous iterations, and the value of $\epsilon$ is chosen to be $10^{-7}$.

In order to evaluate the performance of the registration algorithms, we generate plots (see Fig.~\ref{fig:with_noise}) showing success ratio versus success criteria on rotation error (in degrees)~\footnote{We define success ratio as the number of test cases having rotation error less than success criteria.}. We define the area under the curve in these plots, divided by 180 to normalize between 0 and 1, as AUC.  
AUC expresses a measure of success of registration and so the higher the value of AUC, the better the performance of the network. We measure the misalignment between predicted transformation and ground truth transformation and express it in axis-angle representation and we report the angle as rotation error. As for the translation error, we report the L2 norm of the difference between ground truth and estimated translation vectors. 

\begin{figure}[b!]
\centering
          \includegraphics[width=\linewidth]{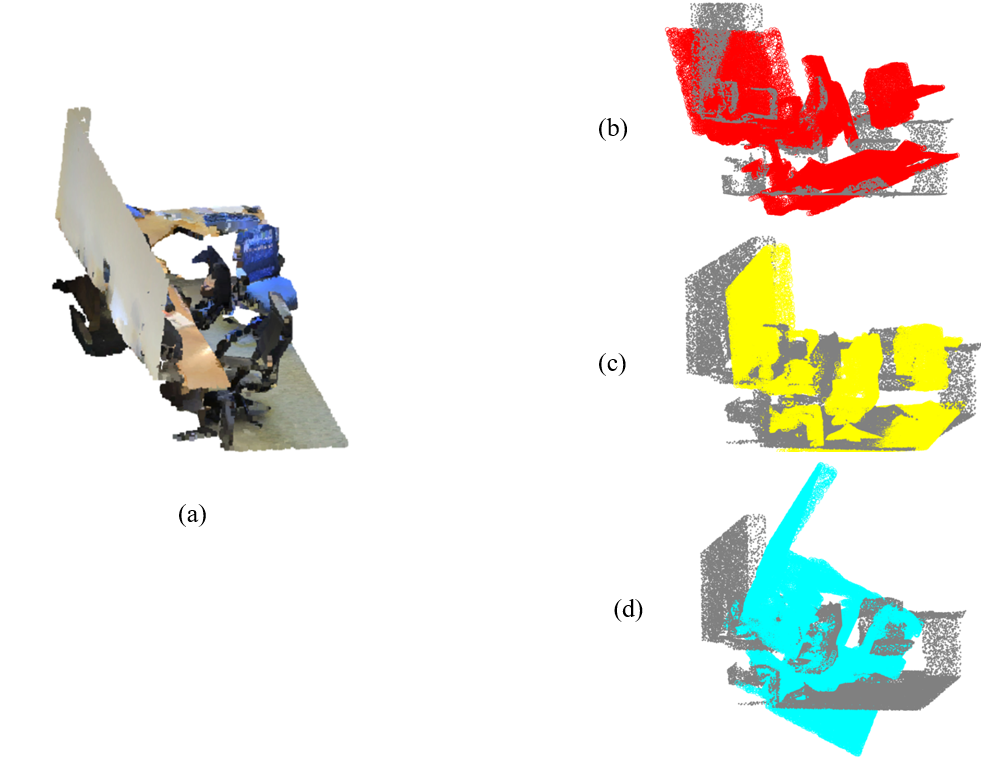}
         \caption{(a) Partial point cloud of office scene from Stanford \emph{S3DIS} indoor dataset (b) red points show registration result of i-PCRNet (19 degrees rotation error) (c) yellow points show registration result of PointNetLK (7 degrees rotation error) (d) cyan points show registration result of 3DSmoothNet (14 degrees rotation error)}
\end{figure}

\subsection{Generalizability versus specificity}
\label{sec:Generalizability}
In the first experiment, i-PCRNet and PointNetLK are trained on 20 different object categories from \emph{ModelNet40} with a total of 5070 models. We perform tests using  100 models chosen from 5 object categories which are not in training data (referred to as unseen categories) with no noise in point clouds. We ensure that same pair of source and template point clouds are used to test all algorithms, for a fair comparison.

We trained i-PCRNet and PointNetLK using multiple object categories and tested them using object categories which are not in training data. There was no noise in source data during training and testing for this experiment. With these tests, we found that AUC for ICP is 0.802, for i-PCRNet is 0.682 and for PointNetLK it is 0.998.

Upon repeating the experiments by training the networks with objects from the same category as the data being tested on, we observe a massive improvement in the AUC for i-PCRNet, going from 0.682 to 0.972. The AUC for ICP and PointNetLK were similar to earlier at 0.862 and 0.998 respectively, and the AUC of PCRNet was 0.998.

These results emphasize that the i-PCRNet and PCRNet, when retrained with object specific information, provide improved registration results compared to ICP as well as the version trained with multiple categories. Their performance is comparable to PointNetLK when trained with object specific information. However, PointNetLK shows better generalization than i-PCRNet across various object categories and has better performance compared to ICP (as also observed by~\cite{aoki2019pointnetlk}). We attribute this to the inherent limitation of the learning capacity of PCRNet to large shape variations, while PointNetLK only has to learn the PointNet representation rather than the task of alignment. However, in the next set of experiments, we demonstrate the definite advantages of PCRNet over PointNetLK and other baselines, especially in the presence of noisy data.

\subsection{Incomplete point cloud}
\label{sec:partial}
\begin{figure}[t!]
    \centering
        \includegraphics[width=0.8\linewidth]{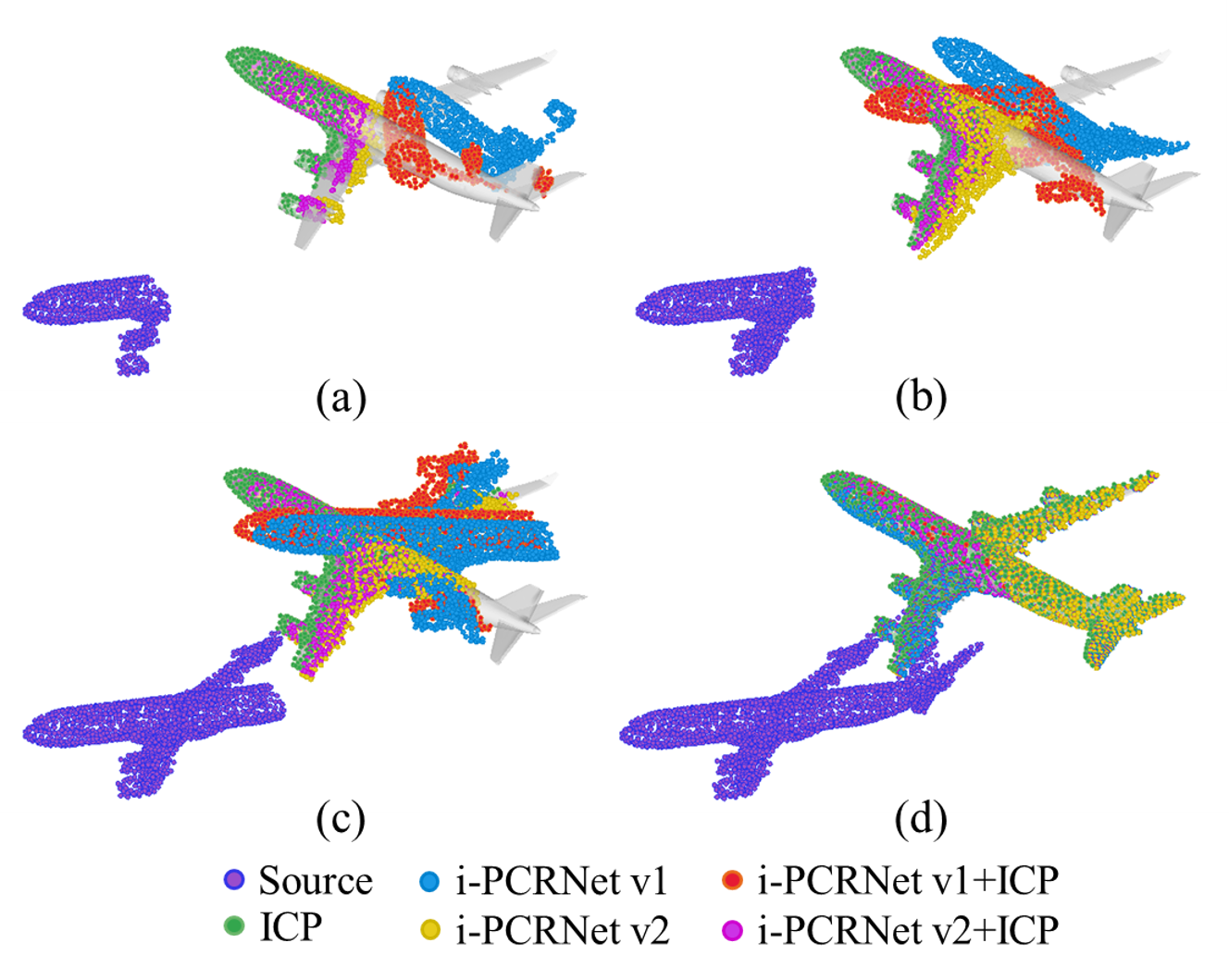}
        \caption{Results for Sec.~\ref{sec:partial}. Rotation and translation error for registering incomplete source point cloud to a template model of airplane. The i-PCRNet v1 and i-PCRNet v2 are trained without and with incomplete source data, respectively. (a) 70\% incompleteness, (b) 50\% incompleteness,  (b) 20\% incompleteness, and (d) complete source data. The performance of i-PCRNet v2 is comparable to ICP (and much better than i-PCRNet v1) even with large amounts of missing points, while being computationally faster than ICP. The ICP refinement produces and improvement only for i-PCRNet v2 and not i-PCRNet v1, since the alignment of i-PCRNet v1 is poor and beyond ICP's capability of refinement. }
    \label{fig:partialAirplane}
    \vspace{-3mm}
\end{figure}

\begin{figure*}[t!]
    \centering
    \begin{subfigure}[b!]{0.32\linewidth}
        \centering
        \includegraphics[width=\linewidth]{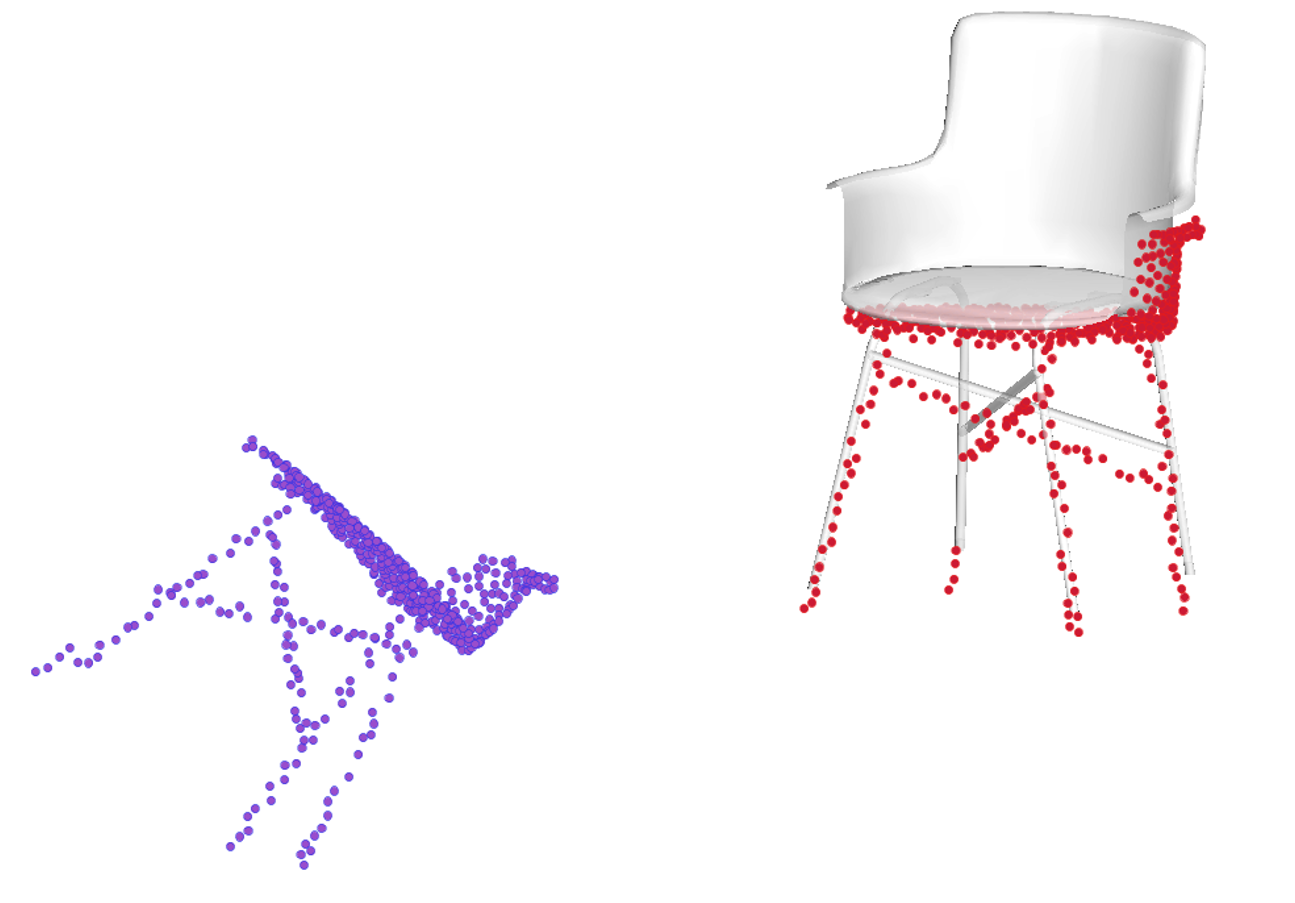}
        \caption{i-PCRNet: Rot error = $4.78^\circ$,\\ Trans error = $0.1035$ units.}
           
    \end{subfigure}
    ~
    \begin{subfigure}[b!]{0.32\linewidth}
        \centering
         \includegraphics[width=\linewidth]{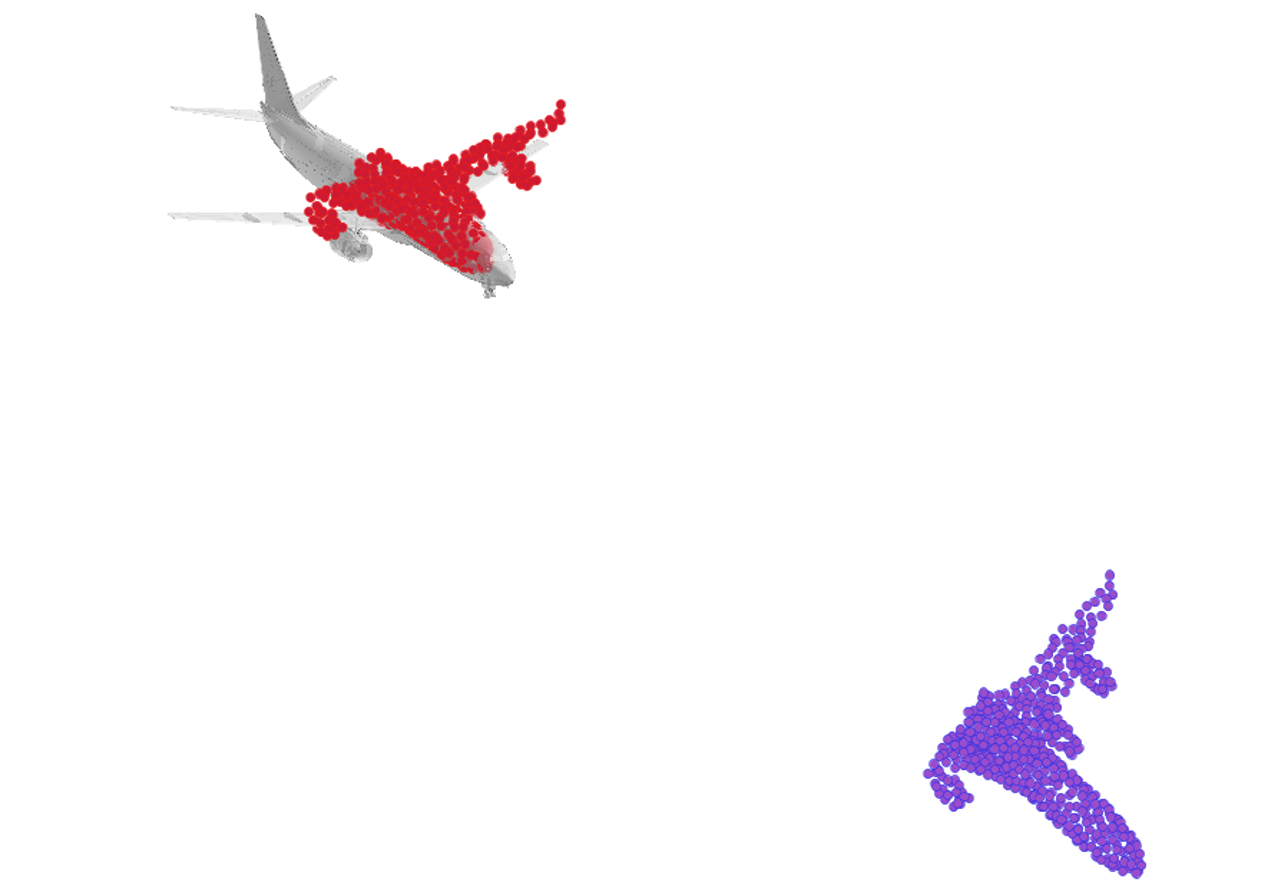}
     \caption{i-PCRNet: Rot error = $4.83^\circ$,\\ Trans error = $0.0952$ units.}
    \end{subfigure}
    ~
    \begin{subfigure}[b!]{0.32\linewidth}
        \centering
        \includegraphics[width=\linewidth]{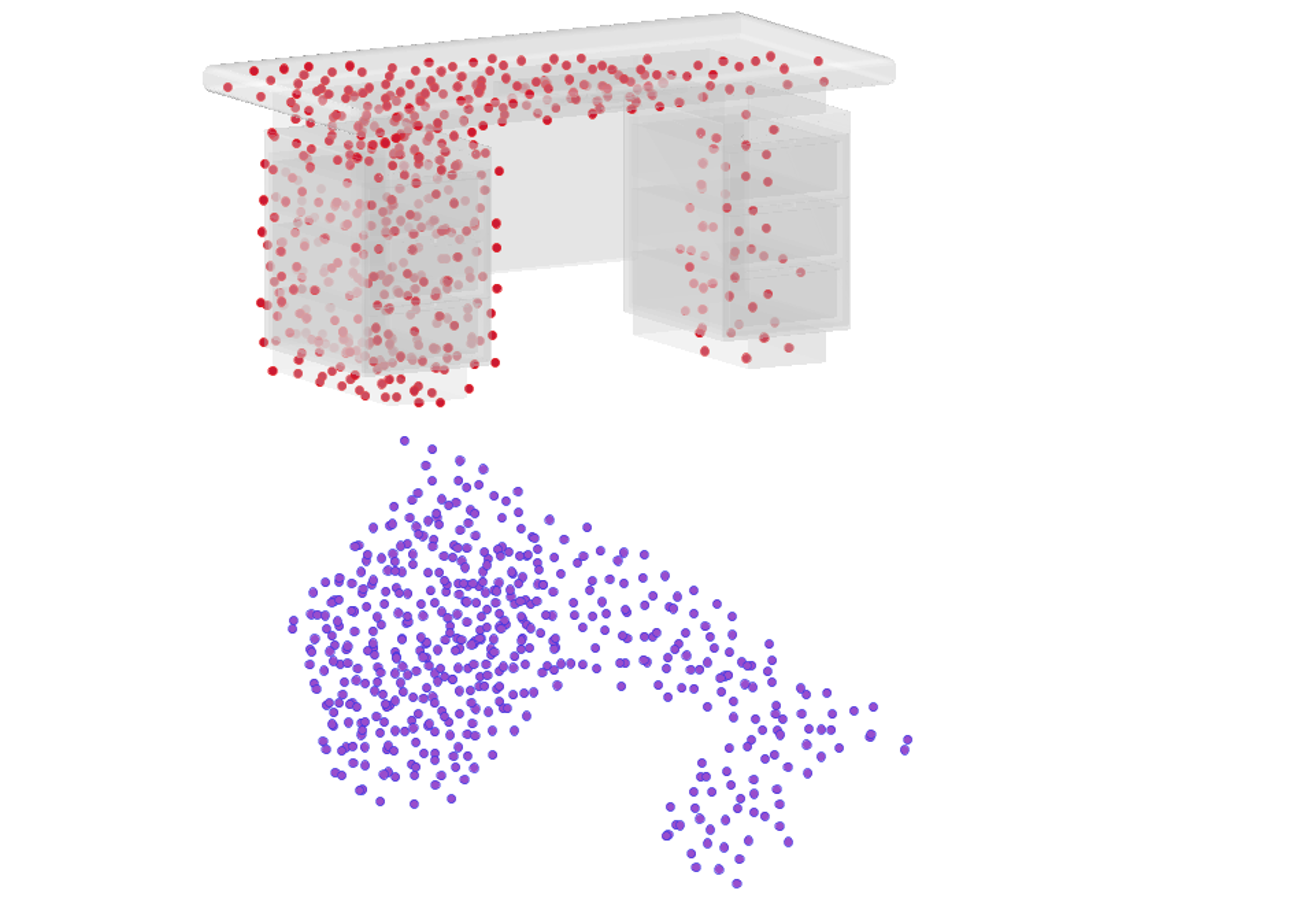}
        \caption{i-PCRNet: Rot error = $7.72^\circ$,\\ Trans error = $0.1673$ units.}
    \end{subfigure}
    ~
    \begin{subfigure}[b!]{0.32\textwidth}
        \centering
         \includegraphics[width=\linewidth]{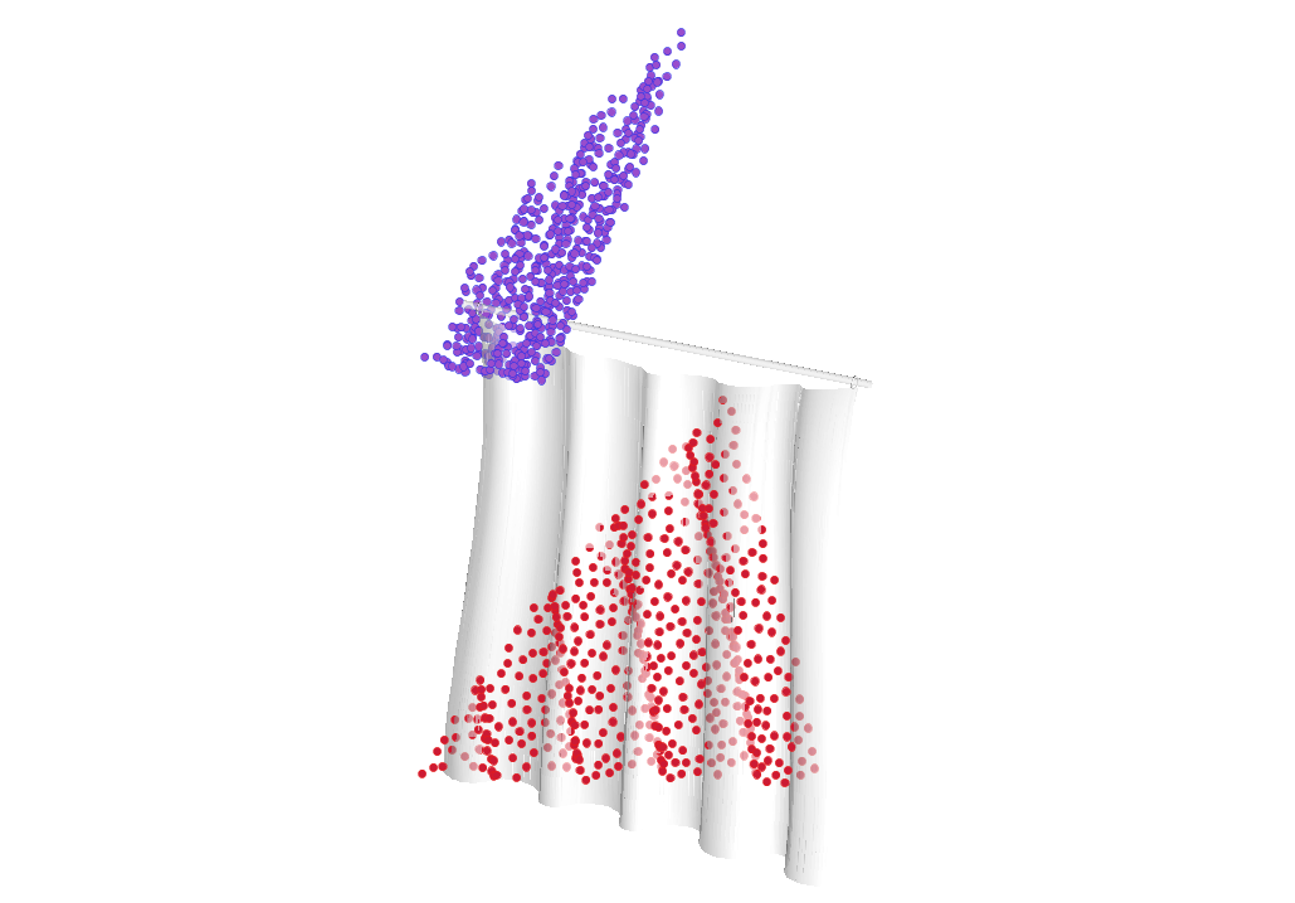}
         \caption{i-PCRNet: Rot error = $13.85^\circ$,\\ Trans error = $0.1339$ units.}
    \end{subfigure}
    ~
    \begin{subfigure}[b!]{0.32\textwidth}
        \centering
         \includegraphics[width=0.9\linewidth]{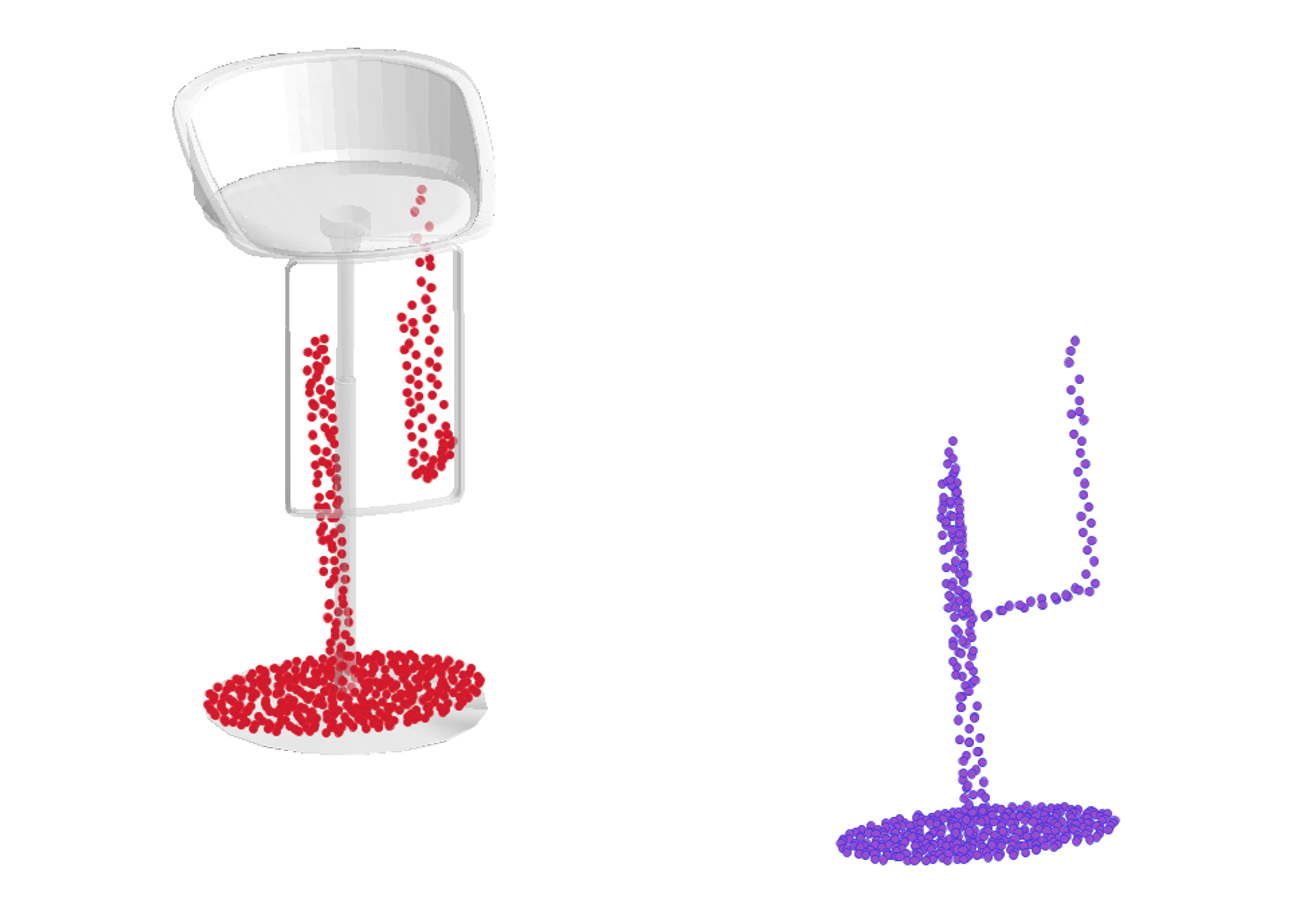}
         \caption{i-PCRNet: Rot error = $6.88^\circ$,\\ Trans error = $0.0466$ units.}
    \end{subfigure}
     ~
    \begin{subfigure}[b!]{0.32\textwidth}
        \centering
         \includegraphics[width=0.9\linewidth]{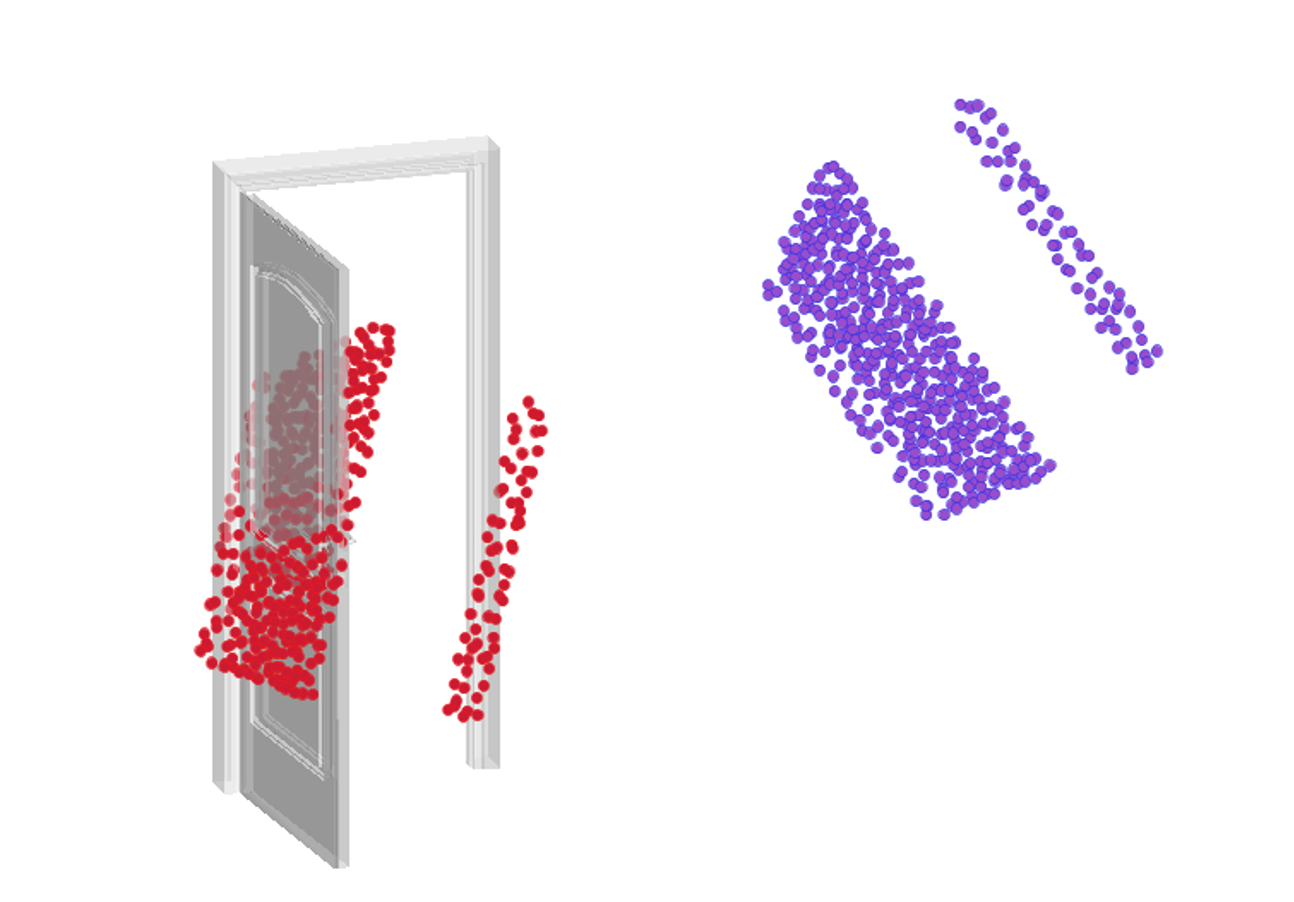}
         \caption{i-PCRNet: Rot error = $21.82^\circ$,\\ Trans error = $0.1937$ units.}
    \end{subfigure}
    
    \caption{Qualitative results for partial point clouds. For each example, template is shown by a grey rendered CAD model, purple points show initial position of source and red points show converged results of i-PCRNet trained on partial data.}
    \label{partial_results}
    \vspace{-3mm}
\end{figure*}

Extending our discussion on robustness when trained with object specific information, we present results for the networks trained on partial source point cloud data. Fig.~\ref{fig:partialAirplane} shows results for varying percentage of incomplete data in the source point cloud. Note that the network trained with partial data is very robust compared to the one that is trained without any partial data. While ICP performs well in all cases, it is computationally slower than iPCRNet (as discussed later in Sec.~\ref{sec:speed}). Further, refining the output of the network with ICP is not always helpful. For instance when the network predicts a wrong pose, ICP refinement can further worsen the alignment as shown in Fig.~\ref{fig:partialAirplane}.
In case of partial source data, i-PCRNet does not perform very well if it hasn’t been trained on the partial data as shown in Fig.{~\ref{partial_results}}.  This might hint at the object-specificity of this approach.

\begin{figure}[h!]
    \centering
    \includegraphics[width=0.8\linewidth, height=5.5cm]{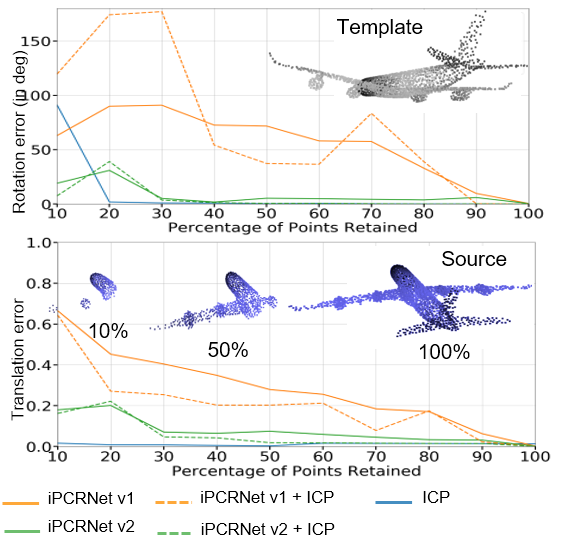}
    \caption{Rotation and translation error for registering incomplete source point cloud to a template model of airplane. The i-PCRNet v1 and i-PCRNet v2 are trained without and with incomplete source data, respectively. The performance of i-PCRNet v2 is comparable to ICP (and much better than i-PCRNet v1) even with large amounts of missing points, while being computationally faster than ICP. The ICP refinement produces and improvement only for i-PCRNet v2 and not i-PCRNet v1, since the alignment of i-PCRNet v1 is poor and beyond ICP's capability of refinement. }
    \vspace{-3mm}
\end{figure}

\begin{figure*}[h!]
    \centering
    \begin{subfigure}[b!]{0.32\linewidth}
        \centering
        \includegraphics[width=\linewidth]{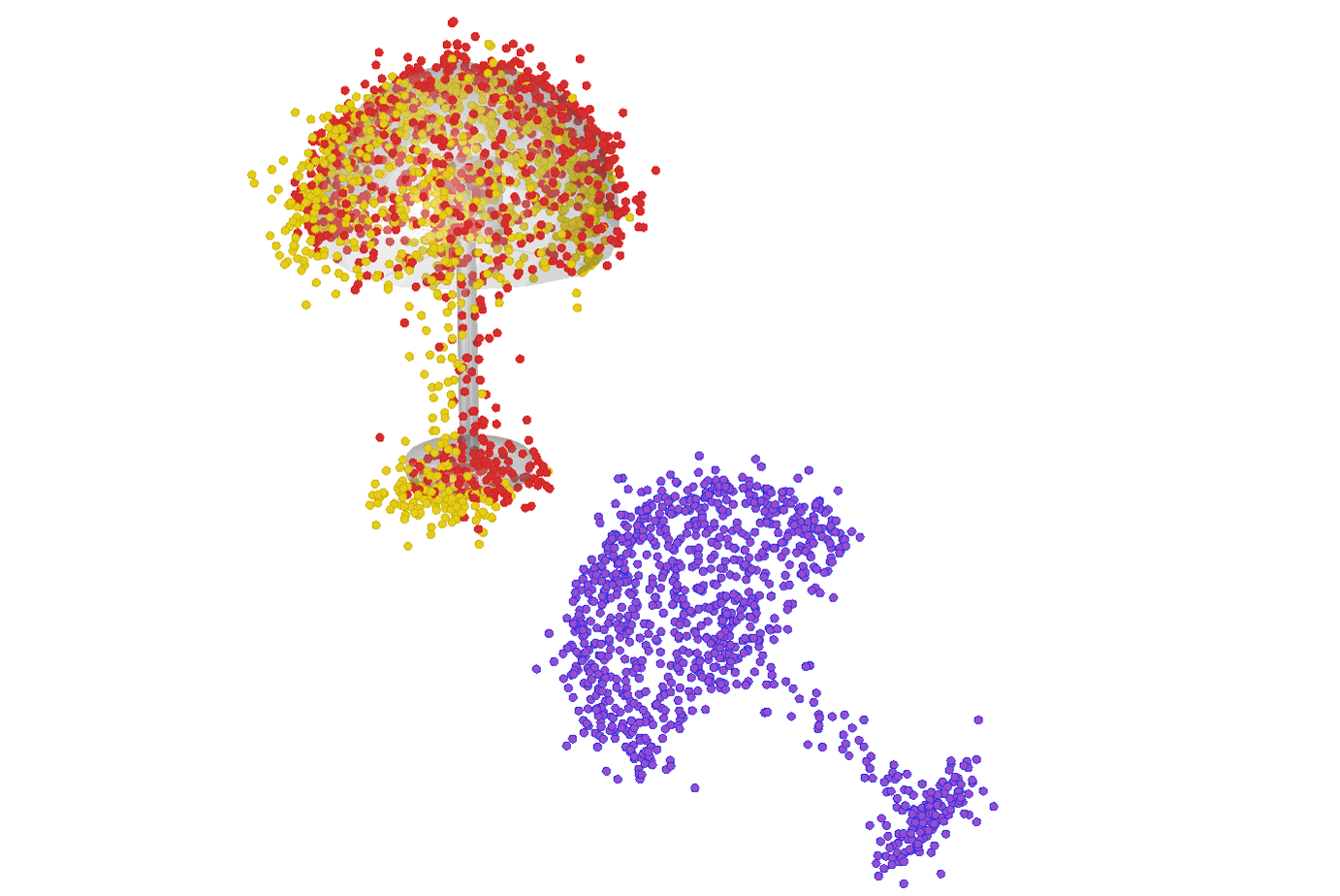}
        \caption{Lamp \\
        i-PCRNet: Rot error = $0.34^\circ$ \\ PointNetLK: Rot error = $11.00^\circ$}
    \end{subfigure}
    \begin{subfigure}[b!]{0.32\textwidth}
        \centering
         \includegraphics[width=\linewidth]{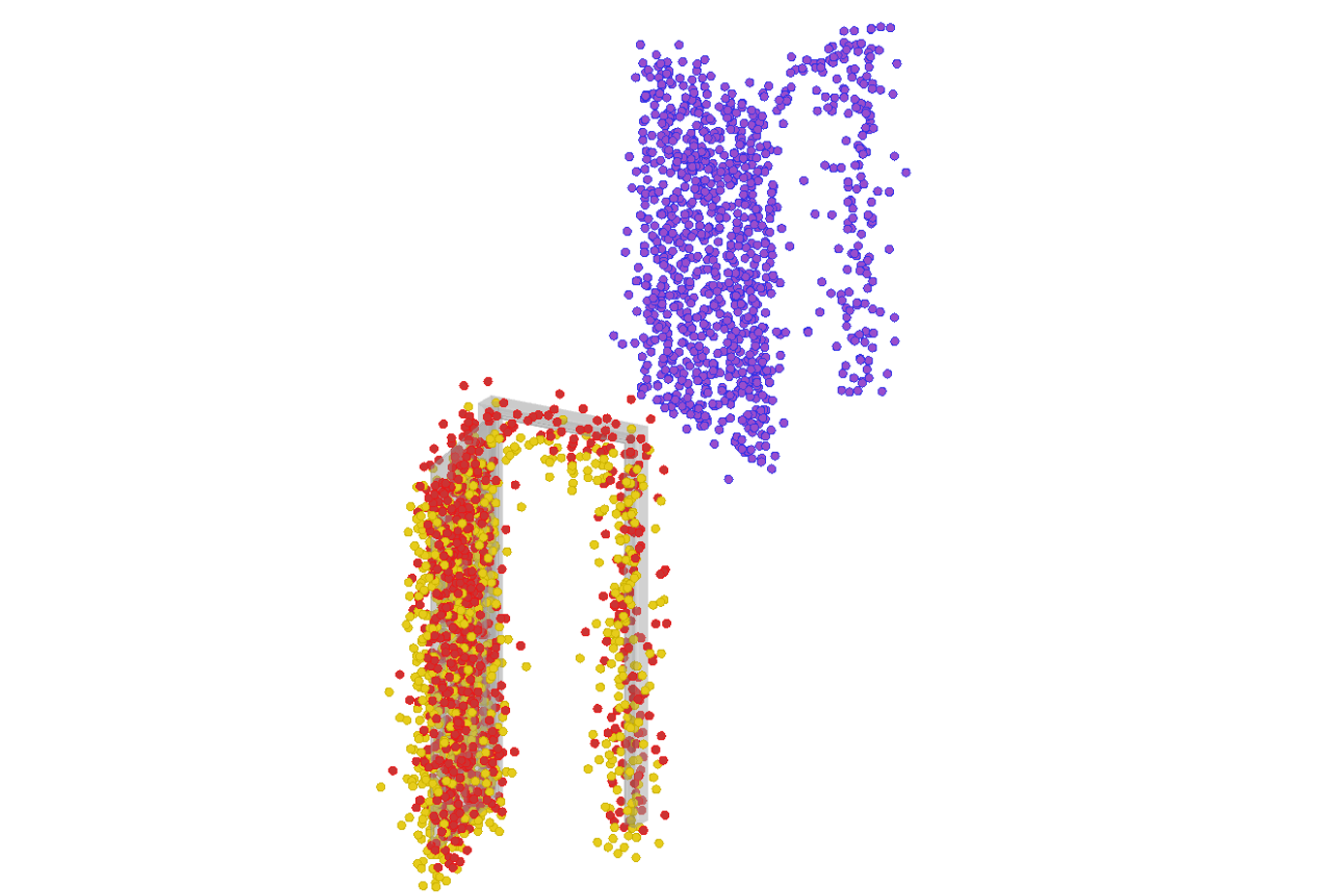}
         \caption{Door \\
         i-PCRNet: Rot error = $0.34^\circ$ \\ PointNetLK: Rot error = $4.73^\circ$}
    \end{subfigure}
    ~
    \begin{subfigure}[b!]{0.32\textwidth}
        \centering
         \includegraphics[width=0.9\linewidth]{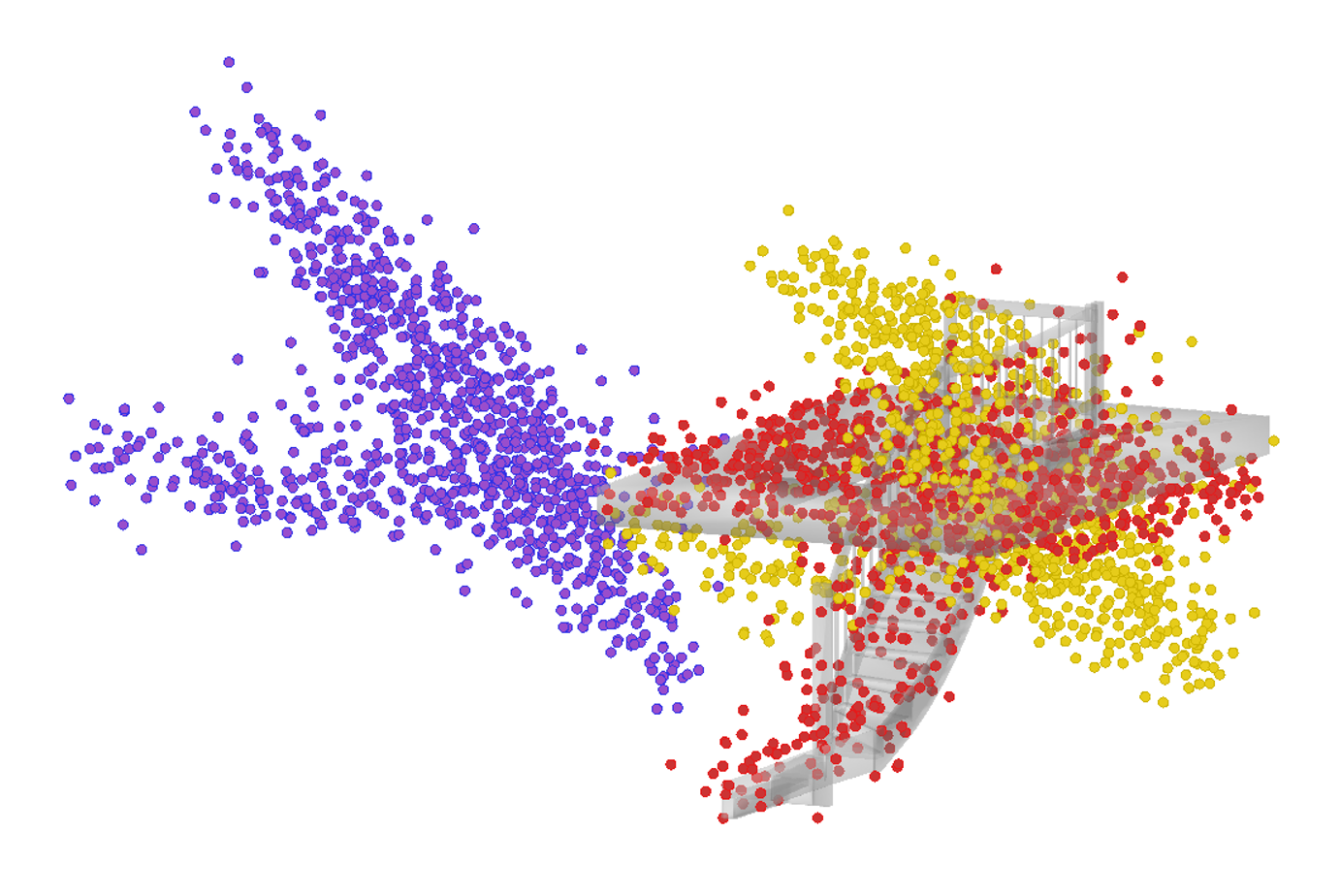}
         \caption{Stairs \\
         i-PCRNet: Rot error = $0.34^\circ$ \\ PointNetLK: Rot error = $51.67^\circ$}
    \end{subfigure}
    ~
    \begin{subfigure}[b!]{0.32\textwidth}
        \centering
         \includegraphics[width=0.9\linewidth]{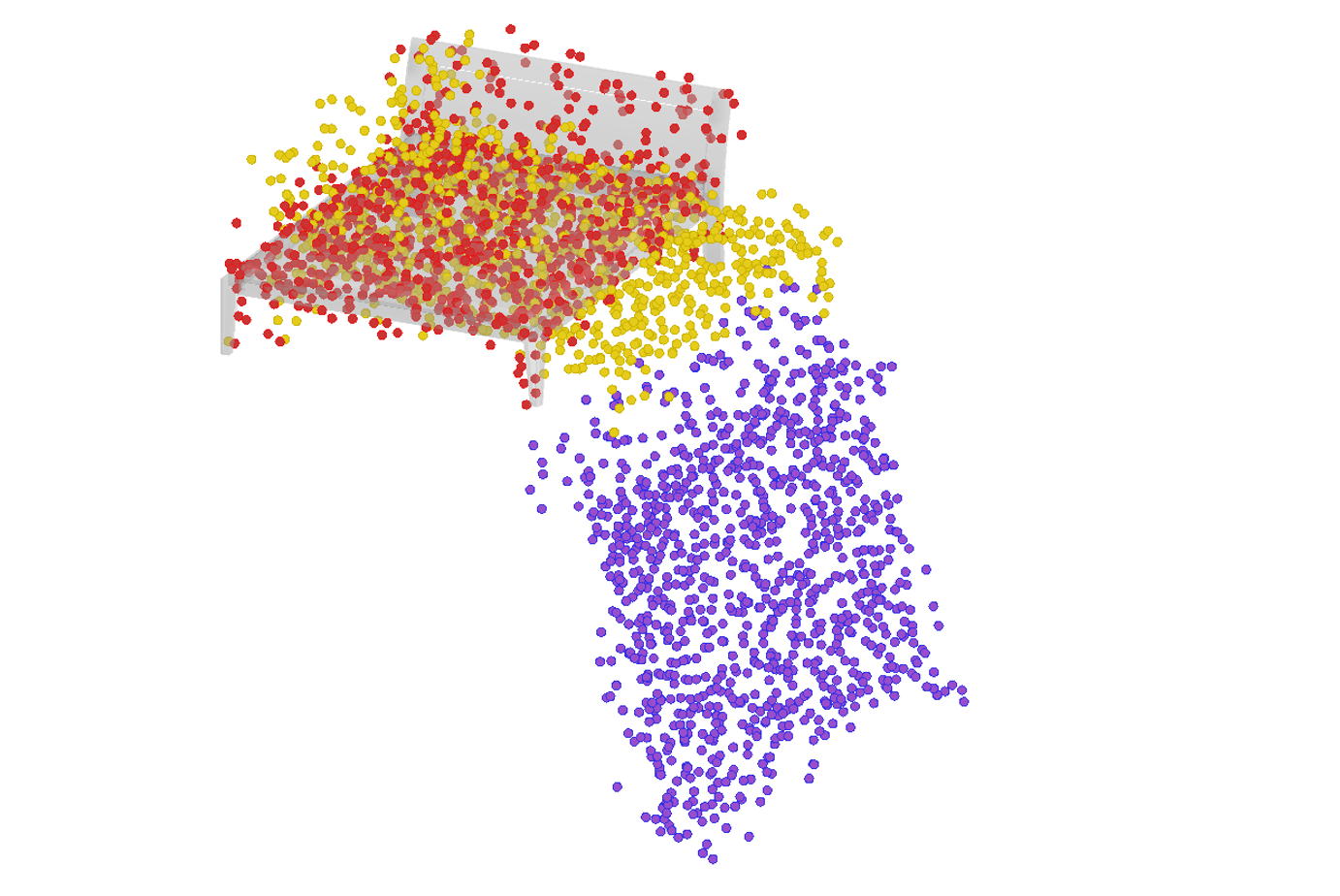}
         \caption{Bed \\
         i-PCRNet: Rot error = $4.15^\circ$,\\ PointNetLK: Rot error = $79.29^\circ$}
    \end{subfigure}
    ~
    \begin{subfigure}[b!]{0.32\textwidth}
        \centering
         \includegraphics[width=0.9\linewidth]{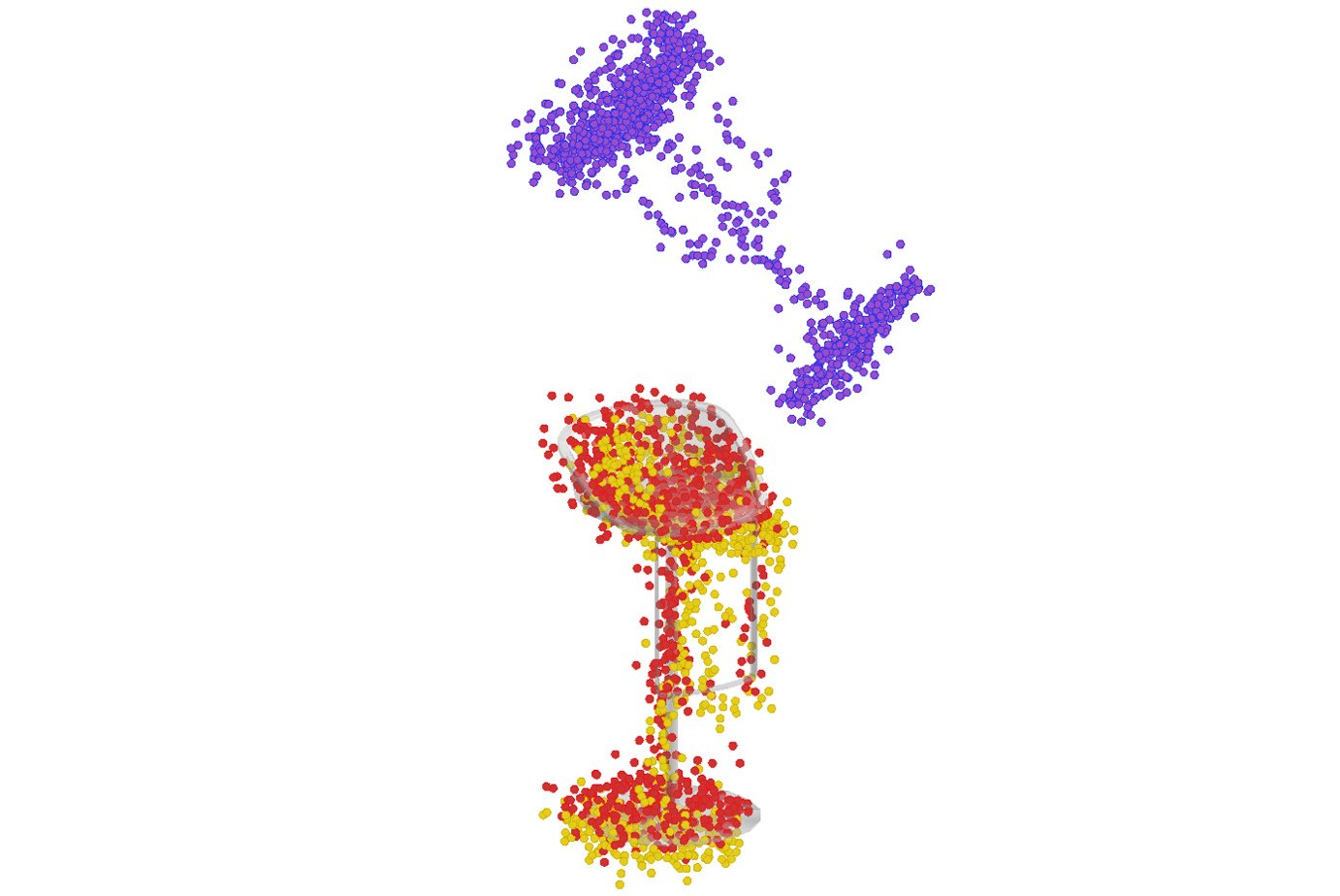}
         \caption{Stool \\
         i-PCRNet: Rot error = $9.13^\circ$ \\ PointNetLK: Rot error = $30.25^\circ$}
    \end{subfigure}
    ~
    \begin{subfigure}[b!]{0.32\textwidth}
        \centering
         \includegraphics[width=0.9\linewidth]{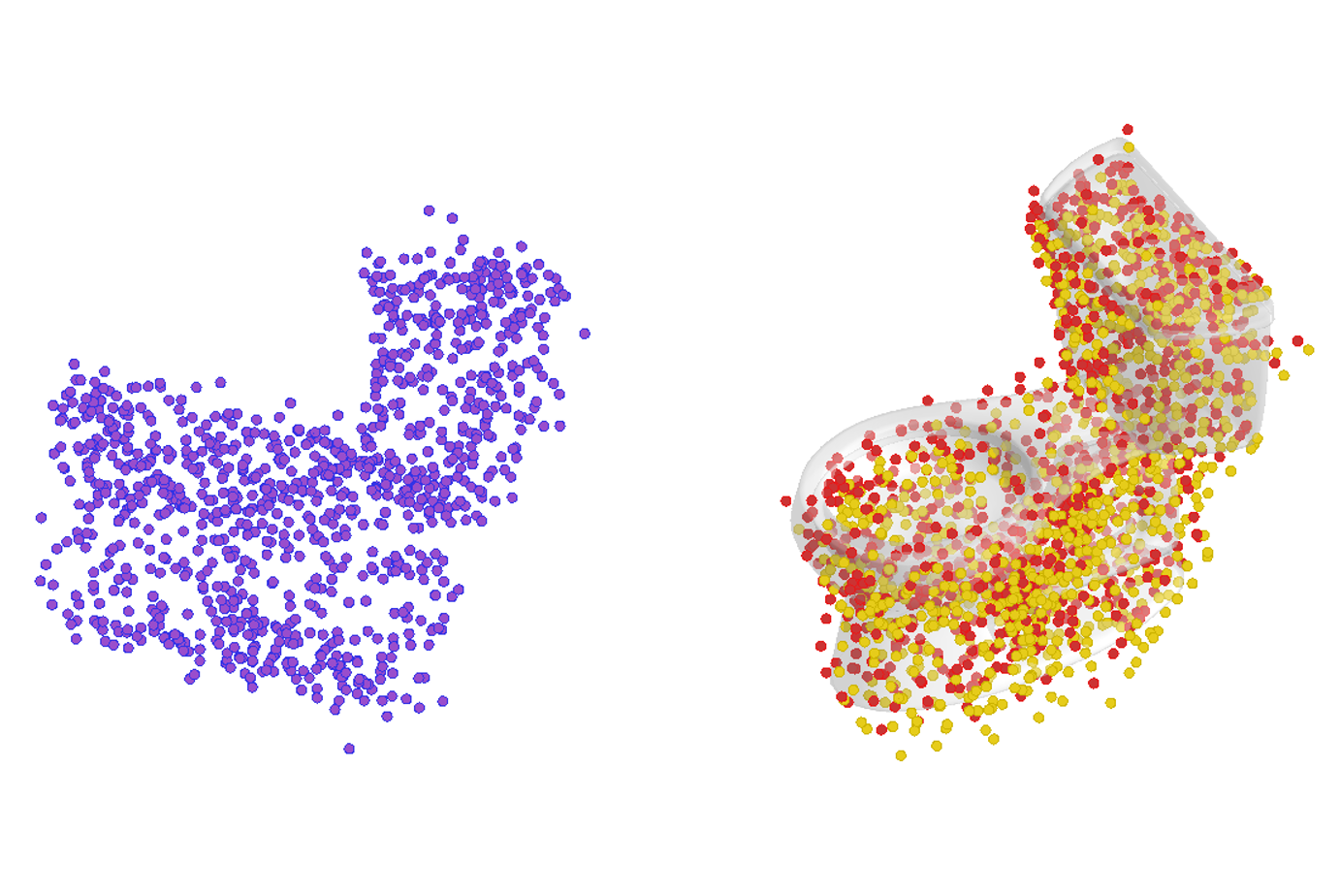}
         \caption{Toilet \\
         i-PCRNet: Rot error = $5.73^\circ$, \\ PointNetLK: Rot error = $8.26^\circ$}
    \end{subfigure}
    \caption{Results for noisy  source point clouds. For each example, the template is visualized by a grey rendered CAD model, purple points show initial position of source, green points show results of ICP, red points show converged results of i-PCRNet trained on data with noise, yellow points show the results of PointNetLK trained on noisy data.}
    \label{fig:noisy_data}
\end{figure*}

\subsection{Gaussian noise}
\label{sec:Gaussian_Noise}

\begin{figure}[ht!]
    \centering
    \includegraphics[width=0.899\columnwidth, height=5.92cm]{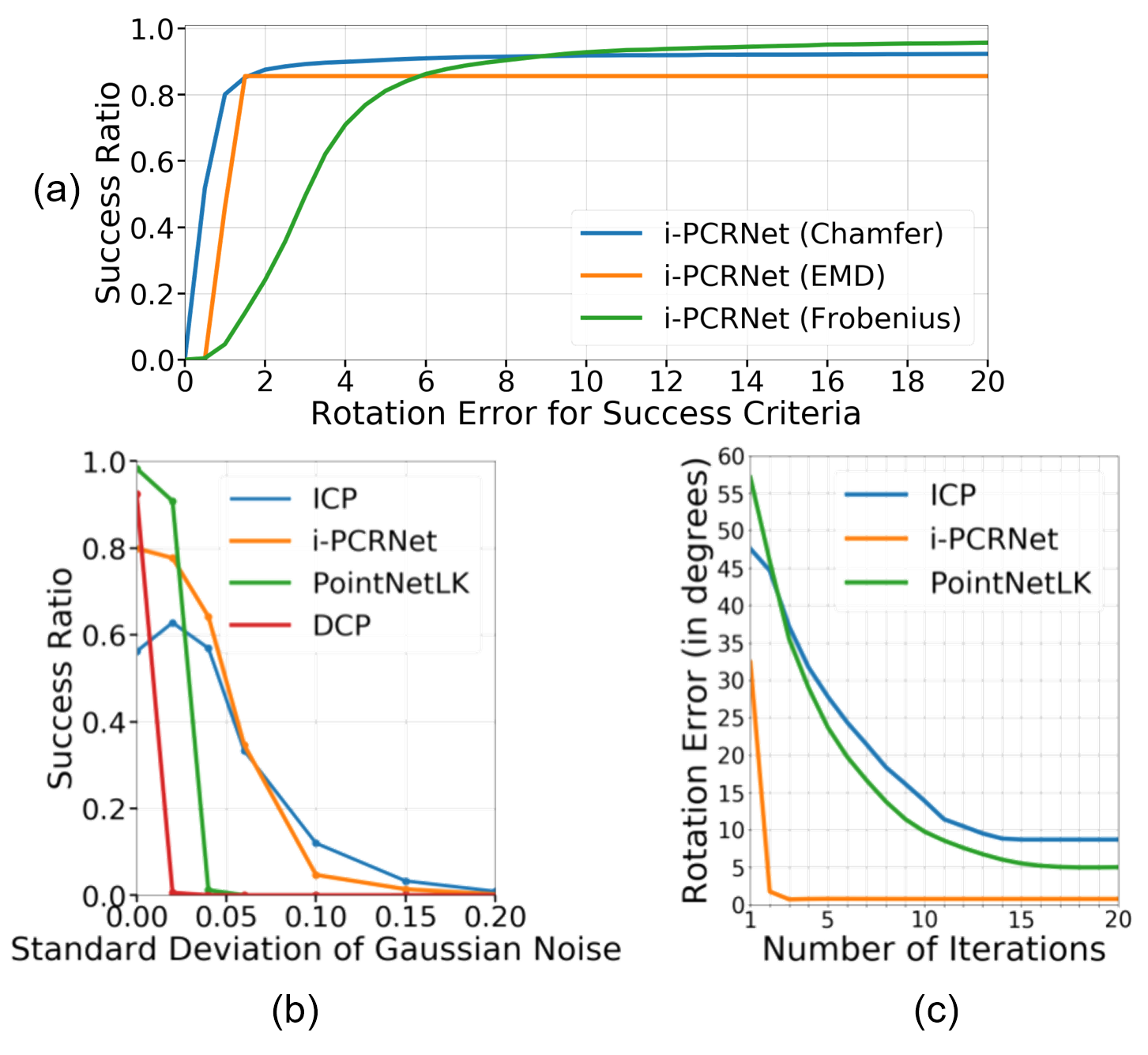}
    \caption{(a) Performance of i-PCRNet for different choices of loss functions. Chamfer loss shows best performance in our experiments. (b) Performance of ICP, i-PCRNet, PointNetLK and DCP for different levels of noise in source point cloud. DCP and PointNetLK are very sensitive to noise and perform the worst. i-PCRNet is very robust to noise in the levels that it has observed during training. (c) Rotation error versus the number of iterations performed to find the pose. i-PCRNet aligns source and template point clouds in least amount of iterations.}
    \label{fig:iterations_vs_error}
    \vspace{-3mm}
\end{figure}

In order to evaluate robustness of our networks to noise, we perform experiments with Gaussian noise in the source points. For our first experiment, we use dataset as described in Sec.~\ref{sec:Generalizability}. We sample noise from a zero mean Gaussian distribution with a standard deviation varying in the range of 0 to 0.04 units. During testing, we compare the methods with noise in source data for each algorithm. We ensured that the dataset has the same pairs of source and template point clouds for a fair comparison. 

For the second experiment, we train the networks only on a specific object category with added Gaussian noise. We test them on the 150 models of the same category with Gaussian noise. In a similar manner, for the third experiment, we train and test the networks on only one noisy model. In all these cases, i-PCRNet is most robust to Gaussian noise, with higher number of successful test cases having smaller rotation error as compared to ICP and PointNetLK (see Fig.~\ref{fig:multi_catg_noise}~\ref{fig:one_catg_noise}~\ref{fig:one_model_noise}). It is worth noting that PointNetLK is very sensitive to noisy data.

Finally, we compare PCRNet that is trained without noise and tested on noisy data, with ICP and PointNetLK. While not being as good as ICP, PCRNet is still competitive, and performs much better than PointNetLK (See Fig.~\ref{fig:one_model_noise_siam}). We present qualitative results in Fig.~\ref{fig:example_registration} using i-PCRNet trained on multiple datasets and testing with noisy data. As expected, the accuracy of i-PCRNet is highest when trained on the same model that it is being tested on. The accuracy drops only a little when trained on multiple models and multiple categories, showing a good generalization as long as there is some representation of the test data in the training. 
Fig.~\ref{fig:noisy_data} shows  an  extension  of  our  discussion where we observe that i-PCRNet performs better than PointNetLK when dealing with noisy data.

Fig.~\ref{fig:iterations_vs_error}(b) shows success ratio versus the amount of noise added to source point clouds during testing. DCP, i-PCRNet and PointNetLK are trained on multiple object categories with Gaussian noise having a maximum standard deviation of 0.04. We observe a sudden drop in the performance of PointNetLK and DCP as the standard deviation for noise increases above 0.02. On the other hand, i-PCRNet performs best in the neighbourhood of the noise range that it was trained on (0.02-0.06), and produces results comparable to ICP beyond that noise level. This shows that i-PCRNet is more robust to noise as compared to PointNetLK. Fig.~\ref{fig:iterations_vs_error}(c) shows the rotation error versus number of iterations in for the different methods. Notice that the i-PCRNet takes only 3 iterations to get close to convergence, compared to the other methods that take upwards of 15 iterations.

\subsection{Computation speed comparisons}
\label{sec:speed}
We use a testing dataset with only one model of car from \emph{ModelNet40} dataset, with Gaussian noise in the source data. We apply 100 randomly chosen transformations with Euler angles in range of $[-45^\circ, 45^\circ]$ and translation values in range of [-1, 1] units. All the networks are trained using multiple models of same category (i.e. car). We compared the performance of i-PCRNet, PCRNet, PointNetLK, DCP, ICP and Go-ICP, as shown in Table~\ref{tb:goicp}. We also develop a variant of i-PCRNet (we refer to this as VoxReg), where the PointNet module is replaced with a VoxNet \cite{maturana2015voxnet}. The comparison methods were chosen to cover a wide spectrum of registration methods, including conventional approaches and learning-based approaches. The learning-based methods use different embeddings such as pointNet, dynamic graph and voxels.

The results demonstrate that Go-ICP converges to a globally optimal solution in all cases with a very small rotation error and translation error, but the time taken is three orders of magnitude more than i-PCRNet and five orders of magnitude more than PCRNet. The VoxReg has an accuracy and computation time similar to ICP. The i-PCRNet has an accuracy similar to Go-ICP, but is computationally much faster, allowing for use in many practical applications. Further, while PCRNet is not as accurate as i-PCRNet, the accuracy may be good enough for a pre-aligning step in applications such as object detection and segmentation~\cite{RW:Wentao}.

\setlength{\tabcolsep}{0.1mm}
\begin{table}[t]
\centering
\caption[m1]{Results from Section \ref{sec:speed}.}
\label{tb:goicp}
\begin{tabular}
                {l|c|c|c|c|c|c|c}\toprule
                 & \multicolumn{2}{c}{Rot. Error} & \multicolumn{2}{c}{Trans. Error} & \multicolumn{2}{c}{Time} & AUC \\
                 & \multicolumn{2}{c}{ (deg)} & \multicolumn{2}{c}{$(\times 10^{-2})$} & \multicolumn{2}{c}{ (ms)} &  \\
                 & $\mu$         & $\sigma$       & $\mu$           & $\sigma$         & $\mu$       & $\sigma$   \\
 \midrule
PCRNet           & 8.82         & 4.82            & 0.77        & 0.08           & 1.89    & 0.39 & 0.95    \\
i-PCRNet & 1.03         & 2.56            & 0.85        & 0.24           & 146      & 30.40   & 0.99   \\
PtNetLK       & 51.80        & 29.63           & 87.83        & 0.54           & 234      & 41.60  & 0.70     \\
ICP             & 11.87        & 31.87           & 2.82        & 3.92           & 407      & 128.0 & 0.93       \\
DCP           & 24.15       & 14.65           & 0.74       & 0.42          & 27.4   & 1.55 & 0.86 \\ 
VoxReg & 13.97       & 10.67           & 5.61       & 3.27          & 459   & 88.4 & 0.92 \\ 
Go-ICP           & 0.45       & 0.19           & 0.16       & 0.07          & 2.7$\times 10^5$   & 1.5$\times 10^5$ & 1.00 \\ 
\bottomrule     
\end{tabular}
\end{table}

\subsection{Sparse Data}
\begin{figure}[h!]
    \centering
    \begin{subfigure}[b]{0.31\linewidth}
        \centering
        \includegraphics[width=\linewidth]{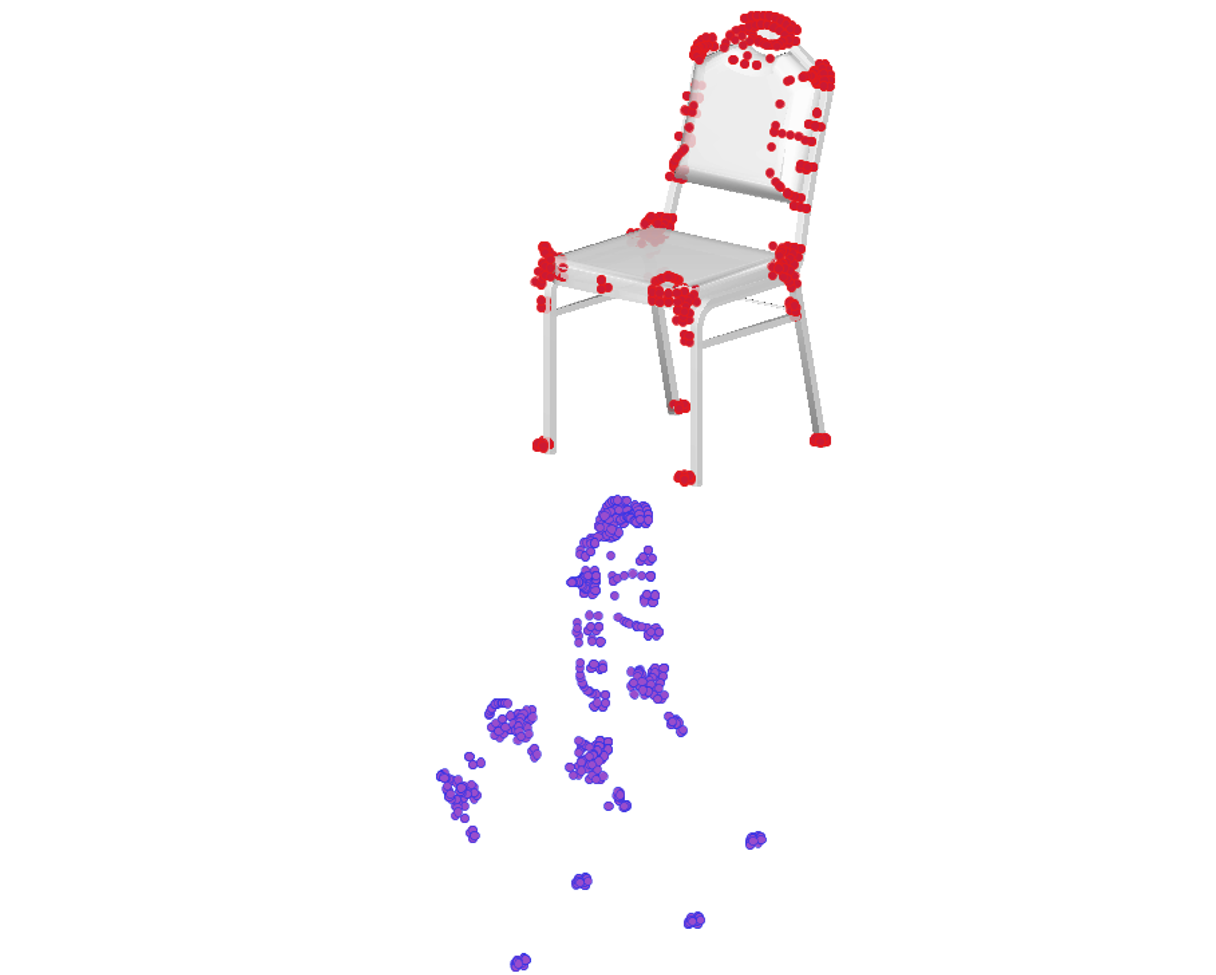}
        \caption{i-PCRNet:\\ Rot error = $3.59^\circ$,\\ Trans error \\= $0.0434$ units.}
    \end{subfigure}
    ~
    \begin{subfigure}[b]{0.31\linewidth}
        \centering
         \includegraphics[width=\linewidth]{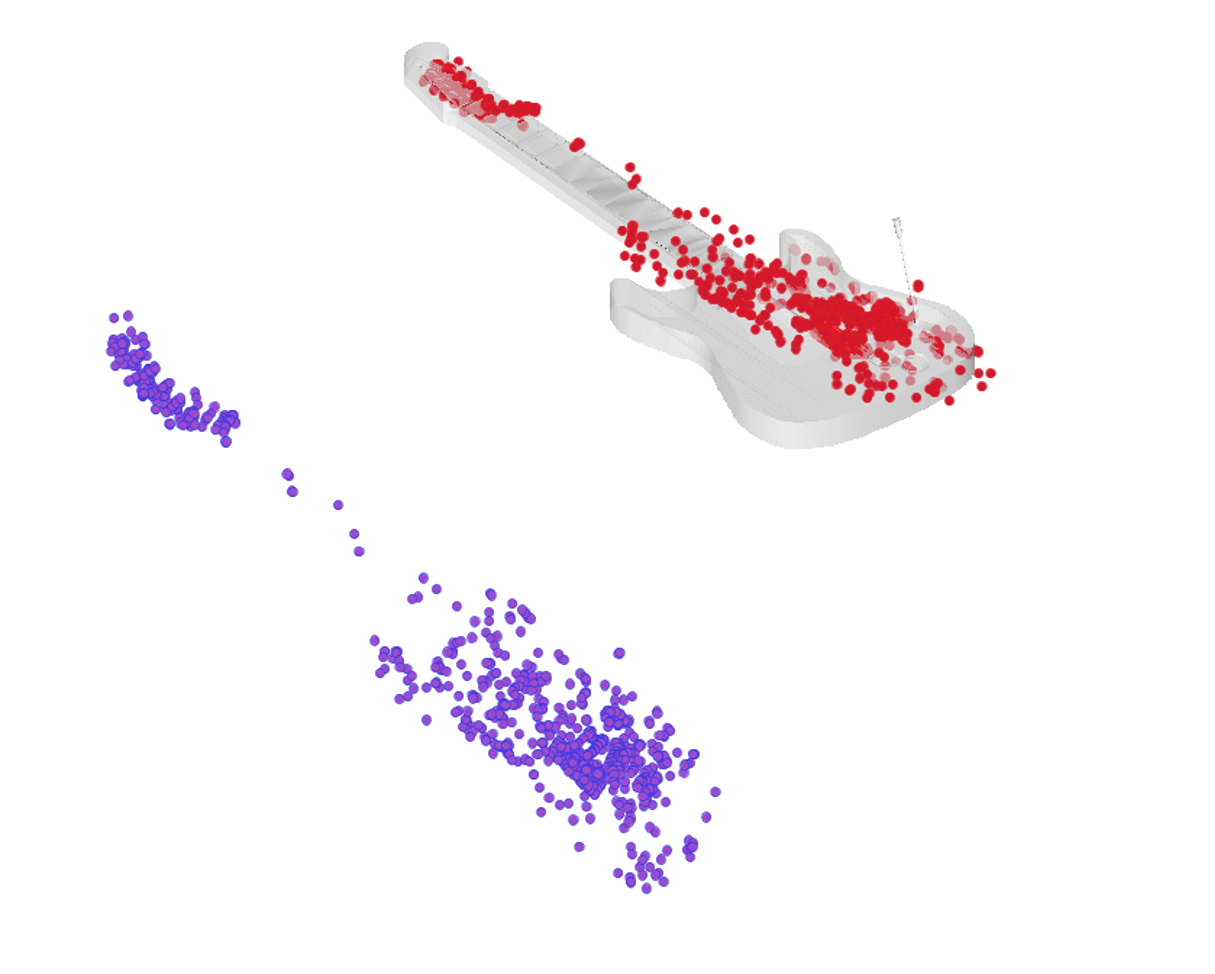}
         \caption{i-PCRNet:\\ Rot error = $25.67^\circ$,\\ Trans error \\= $0.0436$ units.}
    \end{subfigure}
    ~
    \begin{subfigure}[b]{0.31\linewidth}
        \centering
        \includegraphics[width=\linewidth]{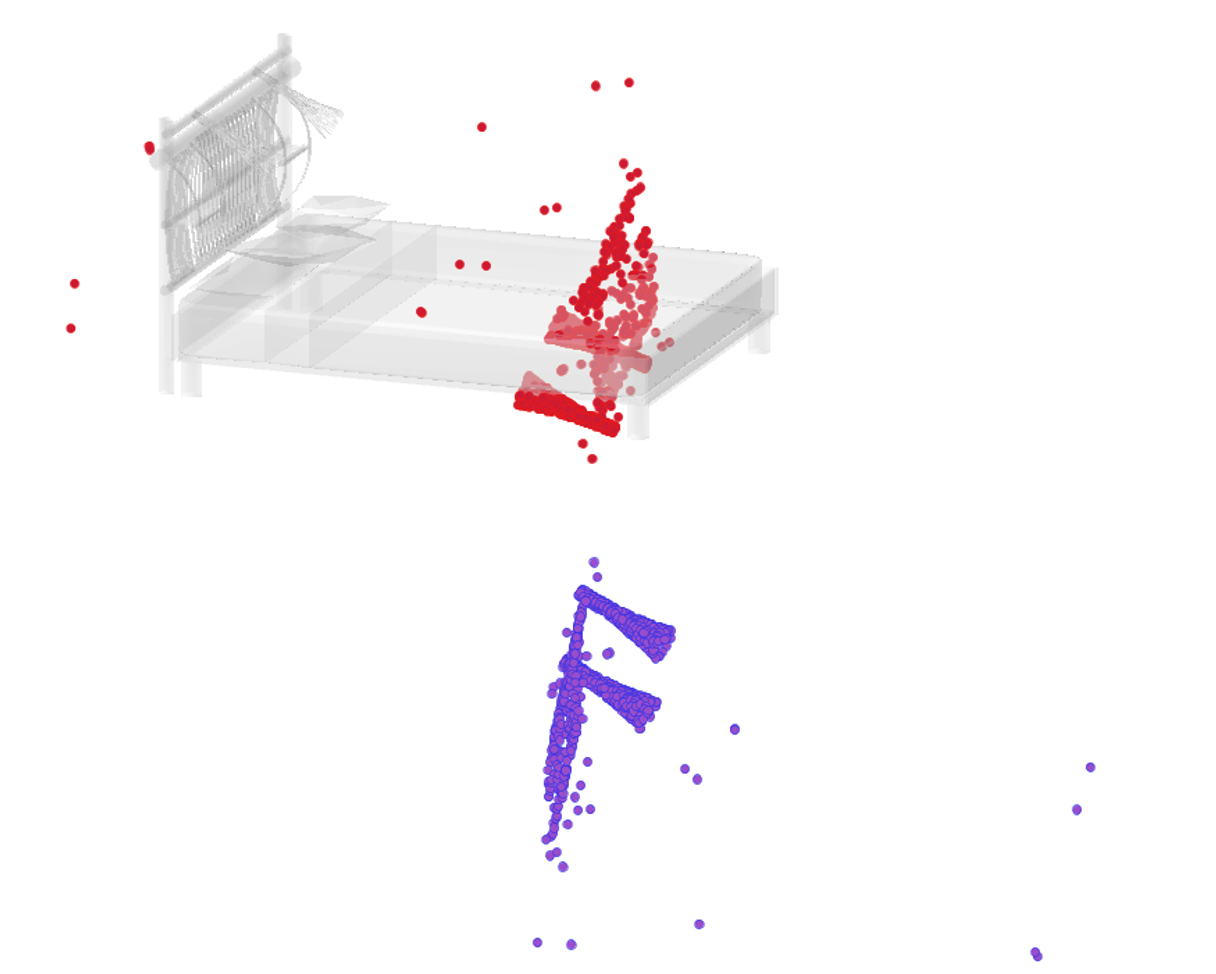}
        \caption{i-PCRNet:\\ Rot error = $173^\circ$,\\ Trans error \\= $0.4867$ units. }
    \end{subfigure}
    \caption{Qualitative results for sparse point clouds. For each example, template is shown by a grey rendered CAD model, purple points show initial position of source and red points show converged results of i-PCRNet trained on sparse point clouds}
    \label{fig:sparse_results}
    \vspace{-3mm}
\end{figure}

We observe from Fig.~\ref{fig:sparse_results} that i-PCRNet trained on sparse data performs better on testing with sparse data.

\section{Model replacement using segmentation}
\label{sec:replacement}
To show qualitative performance on real-world data, we demonstrate the use of i-PCRNet to find the pose and modify the models in an indoor point cloud dataset~\cite{armeni_cvpr16}. We use the semantic segmentation network introduced in PointNet~\cite{qi2017pointnet} to segment a chair from a scene chosen from the Stanford \emph{S3DIS} indoor dataset. We then register it to a chair model from \emph{ModelNet40} dataset using i-PCRNet, which was trained on multiple object categories with noise.

\begin{figure}[t]
    \centering
        \includegraphics[width=0.9\linewidth]{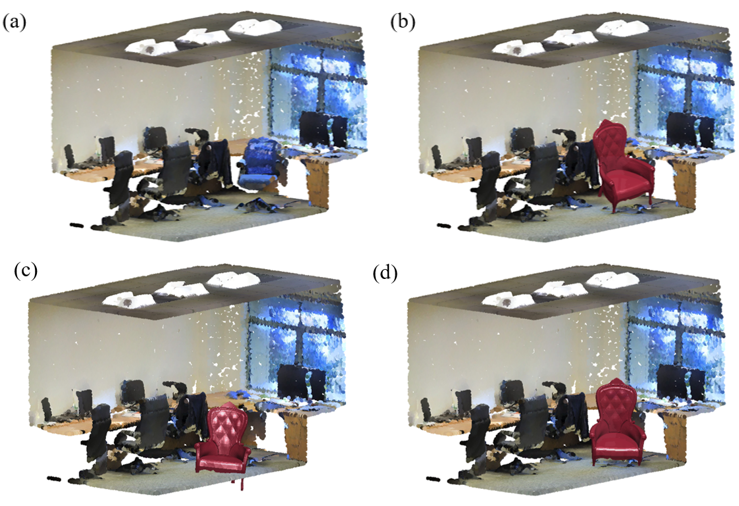}
        \label{fig:room_1}
        \caption{Replacement of chairs in office scene from Stanford \emph{S3DIS} indoor dataset. Red leather chairs shows the replaced chair from \emph{ModelNet40} (a) Original scene. Red leather chair replaced by using registration from (b) ICP, (c) Global registration method, and (d) i-PCRNet. }
    \label{fig:my_label2}
    \vspace{-3mm}
\end{figure}

We replace the original chair with a different chair using the pose obtained from i-PCRNet as shown in Fig.~\ref{fig:my_label2}. Notice that both ICP and global registration method~\cite{izatt2017} fail to register the chair to the right pose, while i-PCRNet accurately registers the point clouds. 


\section{Discussions and future work}
This work presents a novel data-driven framework for performing registration of point clouds using the PointNet representation.

The framework illustrates how data-driven techniques may be used to learn a distribution over appearance variation in point cloud data, including noisy data or category-specificity, and perform better at test time using such a learned prior. The framework can be implemented in an iterative manner to obtain highly accurate estimates comparable to global registration methods. The framework could also be implemented without the iterations, but with  deeper layers to produce two to five orders of magnitude speed improvement compared to popular registration methods.  Finally, this framework also puts into context other recent PointNet-based registration methods in literature such as the PointNetLK. 

Future work would involve integration into larger deep neural network systems, for tasks such as multi-object tracking, style transfer, mapping, etc. Future work may explore the limitations of the learning capacity of the fully-connected registration layers to the size of data distribution. 

\fontsize{9.0}{10.0} \selectfont
{\small
\bibliographystyle{aaai}
\bibliography{aaai.bib}

\begin{thebibliography}{}

\bibitem[\protect\citeauthoryear{Angelina~Uy and
  Hee~Lee}{2018}]{angelina2018pointnetvlad}
Angelina~Uy, M., and Hee~Lee, G.
\newblock 2018.
\newblock Pointnetvlad: Deep point cloud based retrieval for large-scale place
  recognition.
\newblock In {\em Proc. of CVPR},  4470--4479.

\bibitem[\protect\citeauthoryear{Aoki \bgroup et al\mbox.\egroup
  }{2019}]{aoki2019pointnetlk}
Aoki, Y.; Goforth, H.; Srivatsan, R.~A.; and Lucey, S.
\newblock 2019.
\newblock Pointnetlk: Robust \& efficient point cloud registration using
  pointnet.
\newblock In {\em Proc. CVPR},  7163--7172.

\bibitem[\protect\citeauthoryear{Armeni \bgroup et al\mbox.\egroup
  }{2016}]{armeni_cvpr16}
Armeni, I.; Sener, O.; Zamir, A.~R.; Jiang, H.; Brilakis, I.; Fischer, M.; and
  Savarese, S.
\newblock 2016.
\newblock 3d semantic parsing of large-scale indoor spaces.
\newblock In {\em Proceedings of CVPR}.

\bibitem[\protect\citeauthoryear{Baker and Matthews}{2004}]{baker2004lucas}
Baker, S., and Matthews, I.
\newblock 2004.
\newblock {Lucas-Kanade 20 years on: A unifying framework}.
\newblock {\em IJCV} 56(3):221--255.

\bibitem[\protect\citeauthoryear{Besl and McKay}{1992}]{INTRO:ICP}
Besl, P.~J., and McKay, N.~D.
\newblock 1992.
\newblock Method for registration of 3-d shapes.
\newblock In {\em Sensor Fusion IV: Control Paradigms and Data Structures},
  volume 1611,  586--607.
\newblock International Society for Optics and Photonics.

\bibitem[\protect\citeauthoryear{Bristow, Valmadre, and
  Lucey}{2015}]{RW:Semantic}
Bristow, H.; Valmadre, J.; and Lucey, S.
\newblock 2015.
\newblock Dense semantic correspondence where every pixel is a classifier.
\newblock In {\em Proceedings of ICCV},  4024--4031.

\bibitem[\protect\citeauthoryear{Elbaz, Avraham, and
  Fischer}{2017}]{elbaz20173d}
Elbaz, G.; Avraham, T.; and Fischer, A.
\newblock 2017.
\newblock 3d point cloud registration for localization using a deep neural
  network auto-encoder.
\newblock In {\em Proc. of CVPR},  4631--4640.

\bibitem[\protect\citeauthoryear{Eldar \bgroup et al\mbox.\egroup
  }{1997}]{eldar1997farthest}
Eldar, Y.; Lindenbaum, M.; Porat, M.; and Zeevi, Y.~Y.
\newblock 1997.
\newblock The farthest point strategy for progressive image sampling.
\newblock {\em IEEE Transactions on Image Processing} 6(9):1305--1315.

\bibitem[\protect\citeauthoryear{Fan, Su, and Guibas}{2017}]{fan2017point}
Fan, H.; Su, H.; and Guibas, L.~J.
\newblock 2017.
\newblock A point set generation network for 3d object reconstruction from a
  single image.
\newblock In {\em PRoc. of CVPR},  605--613.

\bibitem[\protect\citeauthoryear{Gelfand \bgroup et al\mbox.\egroup
  }{2005}]{gelfand2005robust}
Gelfand, N.; Mitra, N.~J.; Guibas, L.~J.; and Pottmann, H.
\newblock 2005.
\newblock Robust global registration.
\newblock In {\em Symposium on geometry processing}, volume~2, ~5.
\newblock Vienna, Austria.

\bibitem[\protect\citeauthoryear{Georgakis \bgroup et al\mbox.\egroup
  }{2018}]{RW:CAD}
Georgakis, G.; Karanam, S.; Wu, Z.; and Kosecka, J.
\newblock 2018.
\newblock {Matching RGB Images to CAD Models for Object Pose Estimation}.
\newblock {\em arXiv preprint arXiv:1811.07249}.

\bibitem[\protect\citeauthoryear{Gojcic \bgroup et al\mbox.\egroup
  }{2019}]{gojcic2019perfect}
Gojcic, Z.; Zhou, C.; Wegner, J.~D.; and Wieser, A.
\newblock 2019.
\newblock The perfect match: 3d point cloud matching with smoothed densities.
\newblock In {\em Proc. CVPR},  5545--5554.

\bibitem[\protect\citeauthoryear{Guo \bgroup et al\mbox.\egroup
  }{2014}]{guo20143d}
Guo, Y.; Bennamoun, M.; Sohel, F.; Lu, M.; and Wan, J.
\newblock 2014.
\newblock {3D object recognition in cluttered scenes with local surface
  features: a survey}.
\newblock {\em IEEE Transactions on Pattern Analysis and Machine Intelligence}
  36(11):2270--2287.

\bibitem[\protect\citeauthoryear{Held, Thrun, and
  Savarese}{2016}]{held2016learning}
Held, D.; Thrun, S.; and Savarese, S.
\newblock 2016.
\newblock Learning to track at 100 fps with deep regression networks.
\newblock In {\em ECCV},  749--765.
\newblock Springer.

\bibitem[\protect\citeauthoryear{Insafutdinov and
  Dosovitskiy}{2018}]{insafutdinov2018unsupervised}
Insafutdinov, E., and Dosovitskiy, A.
\newblock 2018.
\newblock Unsupervised learning of shape and pose with differentiable point
  clouds.
\newblock In {\em Advances in Neural Information Processing Systems},
  2802--2812.

\bibitem[\protect\citeauthoryear{Izatt, Dai, and Tedrake}{2017}]{izatt2017}
Izatt, G.; Dai, H.; and Tedrake, R.
\newblock 2017.
\newblock Globally optimal object pose estimation in point clouds with
  mixed-integer programming.
\newblock In {\em International Symposium on Robotics Research}.

\bibitem[\protect\citeauthoryear{Li \bgroup et al\mbox.\egroup
  }{2018}]{li2018deepim}
Li, Y.; Wang, G.; Ji, X.; Xiang, Y.; and Fox, D.
\newblock 2018.
\newblock {DeepIM: Deep iterative matching for 6d pose estimation}.
\newblock In {\em Proceedings of ECCV},  683--698.

\bibitem[\protect\citeauthoryear{Liu, Qi, and Guibas}{2019}]{liu2019flownet3d}
Liu, X.; Qi, C.~R.; and Guibas, L.~J.
\newblock 2019.
\newblock Flownet3d: Learning scene flow in 3d point clouds.
\newblock In {\em Proc. CVPR},  529--537.

\bibitem[\protect\citeauthoryear{Lu \bgroup et al\mbox.\egroup
  }{2019}]{lu2019deepicp}
Lu, W.; Wan, G.; Zhou, Y.; Fu, X.; Yuan, P.; and Song, S.
\newblock 2019.
\newblock {DeepICP: An End-to-End Deep Neural Network for 3D Point Cloud
  Registration}.
\newblock {\em arXiv preprint arXiv:1905.04153}.

\bibitem[\protect\citeauthoryear{Lucas and Kanade}{1981}]{lucas1981iterative}
Lucas, B.~D., and Kanade, T.
\newblock 1981.
\newblock An iterative image registration technique with an application to
  stereo vision.
\newblock {\em Proc. of IJCAI}.

\bibitem[\protect\citeauthoryear{Makadia, Patterson, and
  Daniilidis}{2006}]{makadia2006fully}
Makadia, A.; Patterson, A.; and Daniilidis, K.
\newblock 2006.
\newblock {Fully automatic registration of 3D point clouds}.
\newblock In {\em Computer Vision and Pattern Recognition, 2006 IEEE Computer
  Society Conference on}, volume~1,  1297--1304.
\newblock IEEE.

\bibitem[\protect\citeauthoryear{Maron \bgroup et al\mbox.\egroup
  }{2016}]{maron2016point}
Maron, H.; Dym, N.; Kezurer, I.; Kovalsky, S.; and Lipman, Y.
\newblock 2016.
\newblock Point registration via efficient convex relaxation.
\newblock {\em ACM TOG} 35(4):73.

\bibitem[\protect\citeauthoryear{Maturana and
  Scherer}{2015}]{maturana2015voxnet}
Maturana, D., and Scherer, S.
\newblock 2015.
\newblock Voxnet: A 3d convolutional neural network for real-time object
  recognition.
\newblock In {\em IROS},  922--928.

\bibitem[\protect\citeauthoryear{Ovsjanikov \bgroup et al\mbox.\egroup
  }{2010}]{ovsjanikov2010one}
Ovsjanikov, M.; M{\'e}rigot, Q.; M{\'e}moli, F.; and Guibas, L.
\newblock 2010.
\newblock One point isometric matching with the heat kernel.
\newblock {\em Computer Graphics Forum} 29(5):1555--1564.

\bibitem[\protect\citeauthoryear{Pais \bgroup et al\mbox.\egroup
  }{2019}]{pais20193dregnet}
Pais, G.~D.; Miraldo, P.; Ramalingam, S.; Govindu, V.~M.; Nascimento, J.~C.;
  and Chellappa, R.
\newblock 2019.
\newblock {3DRegNet: A Deep Neural Network for 3D Point Registration}.
\newblock {\em arXiv preprint arXiv:1904.01701}.

\bibitem[\protect\citeauthoryear{Qi \bgroup et al\mbox.\egroup
  }{2017a}]{qi2017pointnet}
Qi, C.~R.; Su, H.; Mo, K.; and Guibas, L.~J.
\newblock 2017a.
\newblock Pointnet: Deep learning on point sets for 3d classification and
  segmentation.
\newblock {\em Proc. CVPR} 1(2):4.

\bibitem[\protect\citeauthoryear{Qi \bgroup et al\mbox.\egroup
  }{2017b}]{qi2017pointnet++}
Qi, C.~R.; Yi, L.; Su, H.; and Guibas, L.~J.
\newblock 2017b.
\newblock Pointnet++: Deep hierarchical feature learning on point sets in a
  metric space.
\newblock In {\em Advances in Neural Information Processing Systems},
  5099--5108.

\bibitem[\protect\citeauthoryear{Qi \bgroup et al\mbox.\egroup
  }{2018}]{RW:frustum}
Qi, C.~R.; Liu, W.; Wu, C.; Su, H.; and Guibas, L.~J.
\newblock 2018.
\newblock {Frustum pointnets for 3d object detection from RGB-D data}.
\newblock In {\em Proc. of CVPR},  918--927.

\bibitem[\protect\citeauthoryear{Rusinkiewicz and
  Levoy}{2001}]{rusinkiewicz2001efficient}
Rusinkiewicz, S., and Levoy, M.
\newblock 2001.
\newblock {Efficient variants of the ICP algorithm.}
\newblock In {\em 3dim}, volume~1,  145--152.

\bibitem[\protect\citeauthoryear{Rusu, Blodow, and Beetz}{2009}]{rusu2009fast}
Rusu, R.~B.; Blodow, N.; and Beetz, M.
\newblock 2009.
\newblock {Fast point feature histograms (FPFH) for 3D registration}.
\newblock In {\em ICRA},  3212--3217.

\bibitem[\protect\citeauthoryear{Srivatsan \bgroup et al\mbox.\egroup
  }{2019}]{srivatsan2019registration}
Srivatsan, R.~A.; Zevallos, N.; Vagdargi, P.; and Choset, H.
\newblock 2019.
\newblock Registration with a small number of sparse measurements.
\newblock {\em IJRR}.

\bibitem[\protect\citeauthoryear{Vongkulbhisal \bgroup et al\mbox.\egroup
  }{2018}]{vongkulbhisal2018inverse}
Vongkulbhisal, J.; Irastorza~Ugalde, B.; De~la Torre, F.; and Costeira, J.~P.
\newblock 2018.
\newblock Inverse composition discriminative optimization for point cloud
  registration.
\newblock In {\em Proc. of CVPR},  2993--3001.

\bibitem[\protect\citeauthoryear{Wang and Solomon}{2019}]{wang2019deep}
Wang, Y., and Solomon, J.~M.
\newblock 2019.
\newblock {Deep Closest Point: Learning Representations for Point Cloud
  Registration}.
\newblock {\em arXiv preprint arXiv:1905.03304}.

\bibitem[\protect\citeauthoryear{Wang \bgroup et al\mbox.\egroup
  }{2018}]{wang2018dynamic}
Wang, Y.; Sun, Y.; Liu, Z.; Sarma, S.~E.; Bronstein, M.~M.; and Solomon, J.~M.
\newblock 2018.
\newblock {Dynamic graph CNN for learning on point clouds}.
\newblock {\em arXiv preprint arXiv:1801.07829}.

\bibitem[\protect\citeauthoryear{Wang \bgroup et al\mbox.\egroup
  }{2019}]{wang2019densefusion}
Wang, C.; Xu, D.; Zhu, Y.; Mart{\'\i}n-Mart{\'\i}n, R.; Lu, C.; Fei-Fei, L.;
  and Savarese, S.
\newblock 2019.
\newblock {Densefusion: 6D object pose estimation by iterative dense fusion}.
\newblock In {\em Proc.CVPR},  3343--3352.

\bibitem[\protect\citeauthoryear{Wu \bgroup et al\mbox.\egroup
  }{2015}]{wu20153d}
Wu, Z.; Song, S.; Khosla, A.; Yu, F.; Zhang, L.; Tang, X.; and Xiao, J.
\newblock 2015.
\newblock 3d shapenets: A deep representation for volumetric shapes.
\newblock In {\em Proc. CVPR},  1912--1920.

\bibitem[\protect\citeauthoryear{Xiang \bgroup et al\mbox.\egroup
  }{2018}]{xiang2017posecnn}
Xiang, Y.; Schmidt, T.; Narayanan, V.; and Fox, D.
\newblock 2018.
\newblock {PoseCNN: A Convolutional Neural Network for 6D Object Pose
  Estimation in Cluttered Scenes}.
\newblock In {\em RSS}.

\bibitem[\protect\citeauthoryear{Yang \bgroup et al\mbox.\egroup
  }{2016}]{RW:GO_ICP}
Yang, J.; Li, H.; Campbell, D.; and Jia, Y.
\newblock 2016.
\newblock {Go-ICP: A globally optimal solution to 3D ICP point-set
  registration}.
\newblock {\em IEEE trans. on pattern analysis and machine intelligence}
  38(11):2241--2254.

\bibitem[\protect\citeauthoryear{Yew and Lee}{2018}]{yew20183dfeat}
Yew, Z.~J., and Lee, G.~H.
\newblock 2018.
\newblock 3dfeat-net: Weakly supervised local 3d features for point cloud
  registration.
\newblock In {\em ECCV},  630--646.

\bibitem[\protect\citeauthoryear{Yuan \bgroup et al\mbox.\egroup
  }{2018a}]{RW:Wentao}
Yuan, W.; Held, D.; Mertz, C.; and Hebert, M.
\newblock 2018a.
\newblock Iterative transformer network for 3d point cloud.
\newblock {\em arXiv preprint arXiv:1811.11209}.

\bibitem[\protect\citeauthoryear{Yuan \bgroup et al\mbox.\egroup
  }{2018b}]{yuan2018pcn}
Yuan, W.; Khot, T.; Held, D.; Mertz, C.; and Hebert, M.
\newblock 2018b.
\newblock {PCN: Point Completion Network}.
\newblock In {\em 3DV}.

\bibitem[\protect\citeauthoryear{Zhou and Tuzel}{2018}]{zhou2018voxelnet}
Zhou, Y., and Tuzel, O.
\newblock 2018.
\newblock Voxelnet: End-to-end learning for point cloud based 3d object
  detection.
\newblock In {\em Proc. CVPR},  4490--4499.

\bibitem[\protect\citeauthoryear{Zhou \bgroup et al\mbox.\egroup
  }{2019}]{zhou2019continuity}
Zhou, Y.; Barnes, C.; Lu, J.; Yang, J.; and Li, H.
\newblock 2019.
\newblock On the continuity of rotation representations in neural networks.
\newblock In {\em Proc. CVPR},  5745--5753.

\end{thebibliography}
}

\end{document}